\documentclass{article}

\usepackage{arxiv}

\usepackage[utf8]{inputenc} 
\usepackage[T1]{fontenc}    
\usepackage{hyperref}       
\usepackage{url}            
\usepackage{booktabs}       
\usepackage{amsfonts}       
\usepackage{nicefrac}       
\usepackage{microtype}      
\usepackage{lipsum}		
\usepackage{graphicx}

\usepackage{doi}

\usepackage{cite}
\usepackage{amsmath,amssymb}
\usepackage{algorithmic}

\usepackage{textcomp}
\usepackage{xcolor}

\usepackage{subcaption}
\usepackage{multirow}
\usepackage{caption}

\title{An evaluation of pre-trained models for feature extraction in image classification}

\author{ {Erick da Silva Puls} \\
	Institute of Informatics\\
	UFRGS\\
	Porto Alegre, Brazil \\
	\texttt{00275772@ufrgs.br} \\
	\And
	\href{https://orcid.org/0000-0001-7568-878}{\includegraphics[scale=0.06]{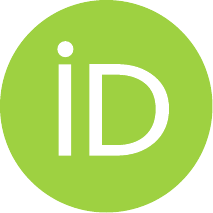}\hspace{1mm}Matheus V. Todescato} \\
	Institute of Informatics\\
	UFRGS\\
    Porto Alegre, Brazil \\
	\texttt{mvtodescato@inf.ufrgs.br} \\
    \And
	\href{https://orcid.org/0000-0002-4499-3601}{\includegraphics[scale=0.06]{orcid.pdf}\hspace{1mm}Joel L. Carbonera} \\
	Institute of Informatics\\
	UFRGS\\
	Porto Alegre, Brazil \\
	\texttt{jlcarbonera@inf.ufrgs.br} \\
}

\date{}

\renewcommand{\headeright}{}
\renewcommand{\undertitle}{}
\renewcommand{\shorttitle}{An evaluation of pre-trained models for feature extraction in image classification}

\hypersetup{
pdftitle={An evaluation of pre-trained models for feature extraction in image classification},
pdfsubject={Image Classification},
pdfauthor={Erick da Silva Puls, Matheus V. Todescato, Joel L. Carbonera},
pdfkeywords={Machine learning, Neural networks, Image classification, Transfer learning, Feature extraction},
}

\begin{document}
\maketitle

\begin{abstract}
	In recent years, we have witnessed a considerable increase in performance in image classification tasks. This performance improvement is mainly due to the adoption of \emph{deep learning techniques}. Generally, deep learning techniques demand a large set of annotated data, making it a challenge when applying it to small datasets. In this scenario, \emph{transfer learning} strategies have become a promising alternative to overcome these issues. This work aims to compare the performance of different pre-trained neural networks for \emph{feature extraction} in image classification tasks. We evaluated 16 different pre-trained models in four image datasets. Our results demonstrate that the best general performance along the datasets was achieved by CLIP-ViT-B and ViT-H-14, where the CLIP-ResNet50 model had similar performance but with less variability. Therefore, our study provides evidence supporting the choice of models for \emph{feature extraction} in image classification tasks.
\end{abstract}


\section{Introduction}

The rapid advancements in technology in the last decades have pushed organizations to produce and accumulate all kinds of data. In the past, critical organizational information was primarily represented by structured data stored in databases. However, nowadays a significant part of this information is represented in an unstructured way, for instance as images \cite{pferd:10}.

There is a need to develop approaches capable of recovering and evaluating images in applications of several fields \cite{pferd:10}. In that sense, one of the challenges concerning image recovery is that the semantic content of images is not apparent, so this information is not easily acquired through direct queries. An alternative for recovering images is annotating them first \cite{hollink:03} in a way that allows us to retrieve them by querying for the annotations. However, it is necessary to bear in mind that manual annotation of large databases of images is time-consuming and impractical. In this context, machine learning can be used to automatically classify these large databases of images, thus enabling retrieval through direct queries.

Image classification (IC) tasks aim to classify the image as a whole by assigning a specific label to it. Usually, labels in an IC task refer to objects that appear in the image, kinds of images (photographs, drawings, etc.), feelings (sadness, happiness, etc.), etc \cite{lanchantin:21}. 

Most of the recent approaches for IC are based on deep neural network (DNN) architectures. These architectures usually demand a large set of annotated data, making it challenging to apply \emph{deep learning} when small amounts of data are available. In this scenario, \emph{transfer learning} strategies have become a promising alternative to overcome these issues. One of the main alternatives of \emph{transfer learning} is through \emph{feature extraction}, where models that were trained on large datasets can be used for producing informative features that can be used by another classifier. By using \emph{transfer learning}, we can leverage knowledge previously learned by neural network models on a large dataset and use this knowledge in a context where just small datasets are available.

There are currently several large datasets available, such as Imagenet \cite{deng:09}, and a range of models that were pre-trained on these datasets \footnote{Some pre-trained models can be found in \url{https://pytorch.org/vision/stable/models.html}}. The literature suggests that particular tasks on distinct datasets can benefit from different pre-trained models \cite{mallouh:19,arslan:21}. 

It is important to notice that there are different approaches for transfer learning for image classification, such as \emph{fine-tuning} and \emph{feature extraction}. When adopting \emph{fine-tuning}, a neural network that was pre-trained in a big dataset is retrained in a novel task, for which usually only a small dataset is available. The goal, in this approach, is to use the knowledge (represented by the weights of the model) acquired in the first training process as a starting point for the training in the second task, and the weights of the pre-trained model are updated during the training in the target task. In the case of \emph{feature extraction}, the pre-trained model is used just for extracting features that represent the images and that can be used as input for a classifier. Notice that in this approach the pre-trained model is kept frozen, that is, their weights are not updated during the training of the classifier used in the target task. Some studies \cite{kieffer2017convolutional, mormont2018comparison} comparing fine-tuning and feature extraction demonstrate that fine-tuning achieves higher performance, but the results also suggest that feature extraction achieves a comparable performance while requiring fewer computational resources for training. In this context, the main goal of this work is to compare and evaluate the performance of \emph{feature extraction} (FE) of various pre-trained models in the task image classification. 

In this study, Geological Images \cite{TodescatoGBC23,abel2019knowledge,todescato2024multiscale}, Stanford Cars \cite{KrauseStarkDengFei:13}, CIFAR-10 \cite{krizhevsky:09}, and STL10 \cite{coates:11} are the datasets adopted for analyzing the performance of FE of the following pre-trained models: AlexNet, ConvNeXt Large, DenseNet-161, GoogLeNet, Inception V3, MNASNet 1.3, MobileNet V3 Large, RegNetY-3.2GF, ResNeXt101-64x4D, ShuffleNet V2 X2.0, SqueezeNet 1.1, VGG19 BN, VisionTransformer-H/14, Wide ResNet-101-2, and both CLIP-ResNet50, and CLIP-ViT-B. We evaluate the performance of the considered pre-trained models using these metrics: accuracy, macro F1-measure, and weighted F1-measure.

Our results indicate that the pre-trained models CLIP-ResNet50, CLIP-ViT-B, and VisionTransformer-H/14 had significantly better performance than the other considered pre-trained models for all datasets. It is important to notice that these are the only three models among those considered in our experiments that include transformers in their architecture, while the others are based solely on CNN architectures. Our analysis also indicates that there are differences regarding the pattern of performances of these three transformer-based architectures in comparison with the performances of the CNN-based architectures across the datasets, in all the considered metrics. These differences become evident when we analyze the Pearson correlation in Section \ref{sec:results}. Moreover, our analysis suggests that the Stanford Cars dataset is the most challenging of all datasets analyzed. We hypothesize that it is due to its large number of classes, few samples per class, and the inclusion of images with different sizes and features at different scales.

The remainder of this paper is structured as the following. In Section \ref{sec:rel_works} we discuss the related work. In Section \ref{sec:experiments} we present our experiments and discuss our results. Finally, Section \ref{sec:conclusion} presents the conclusions.

\section{Related Works}\label{sec:rel_works}

The TL approach based on FE has been adopted for IC in several domains, such as Biomedicine \cite{alzubaidi:21} and Geology \cite{dosovitskiy:20,maniar:18,karpatne:18}. In this work, we reviewed the literature covering the last five years and focused on comparing the performance of FE for different pre-trained models. In total, we selected eighteen papers for our literature overview, of which five are specific to IC in Geology. In our literature review, twenty-three different models were found, among which the most recurrent pre-trained models were the VGG16 and the Inception V3, both for studies in general and those specific to Geology, and the third most used model was the AlexNet. Of the eighteen articles, only two did not use the ImageNet dataset for pre-training the models. Therefore, we can observe that ImageNet is one of the most used datasets for pre-training models for IC.


There is a range of pre-trained models that can be applied for transfer learning. The main expected result of FE from the pre-trained models is to improve classification quality. The size and similarity of the target dataset and the source task can be used to choose the pre-trained model \cite{fawaz:18}. 

The literature suggests that each dataset may need a different pre-trained model. For instance, for plankton classification \cite{luminiAndnanni:19}, when adopted as a feature extractor the Inception V3, AlexNet, VGG16, VGG19, ResNet50, ResNet101, DenseNet-161, and GoogLeNet pre-trained models with three different datasets, the DenseNet-161 model presented the best results. 
On the other hand, when classifying pathological brain images, Kaur \& Gandhi (2020) found that among eight pre-trained models, the AlexNet showed the best results. 

Finally, using the CIFAR-10 dataset, and experimenting with the Inception V3, GoogLeNet, SqueezeNet 1.1, and DarkNet53, ShuffleNet models, \citeonline{kumarAndanuarAndhafizah:22} found that, overall, the Inception V3 model achieved the highest accuracy, as well as higher values in other evaluation metrics including precision, sensitivity, specificity, and F1-score \cite{kumarAndanuarAndhafizah:22}. 

In summary, for specific application needs, finding the right pre-trained model can be challenging. Different models can present better results for different datasets and different parameters of performance \cite{aboubAndzengelerAndhandmann:22}. Therefore, it is essential to systematically investigate the usability of several pre-trained models to find the best match for specific datasets.


Although ML has been successfully applied in Geosciences in the last years, TL for IC in this domain is still not as much explored as in other areas \cite{delima:19}.

Recent papers are showing the capacity of DL and TL to facilitate the analysis of uninterpreted images that have been neglected due to a limited number of experts, such as fossil images, slabbed cores, or petrographic thin sections \cite{delima:19}, or even for environmental images \cite{sun:21}. The ability to create distinctive models for specific sets of data allows a versatile application of those techniques. 

When comparing pre-trained models, de Lima and colleagues (2019) found that both MobileNet V2 and Inception V3 showed promising results on geologic data interpretation, with MobileNet V2 having slightly better results. Also, when Sun and colleagues (2021) compared the performance of AlexNet, VGG16, ResNet50, GLNet (AlexNet), GLNet (VGG16), and GLNet (ResNet) pre-trained models on remote sensing scene classification using FE, the authors found that their proposed new model shows better results compared to other traditional DNN architectures. The proposed model GLNet (VGG16), which uses VGG16 as its base, got over 95\% accuracy in analyzing a clear environment and over 94\% in a cloudy environment. In contrast, the traditional VGG16 got over 93\% and over 78\%, for clear and cloud environments, respectively \cite{sun:21}.

On seismic imaging classification, which is essential for oil and gas exploration, \citeonline{chevitarese:18}, applied the TL of FE to train a DNN and found improvements in accuracy \cite{chevitarese:18,chevitarese:18efficient}. Finally, \citeonline{cunha:20} also trained a CNN specially developed for the fault identification problem using a dataset with synthetic patches whose seismic signal frequency content does not match that of real data. They used the first layers of this pre-trained CNN as a feature extractor to classify another dataset \cite{cunha:20}.

Table \ref{tab:Table2} presents a summary of the literature review, representing the pre-trained models used by each study identified in our review process.
Table \ref{tab:Table3} presents the target datasets used in each study for comparing the performance of the feature extractors.

Table \ref{tab:Table2} provides a summary of the reviewed literature, emphasizing which pre-trained models were used for FE in each study considered in our review.

\begin{table*}[!ht]
\centering
\caption{Summary of literature review - pre-trained models}
\label{tab:Table2}
\includegraphics[scale=.45, angle=90]{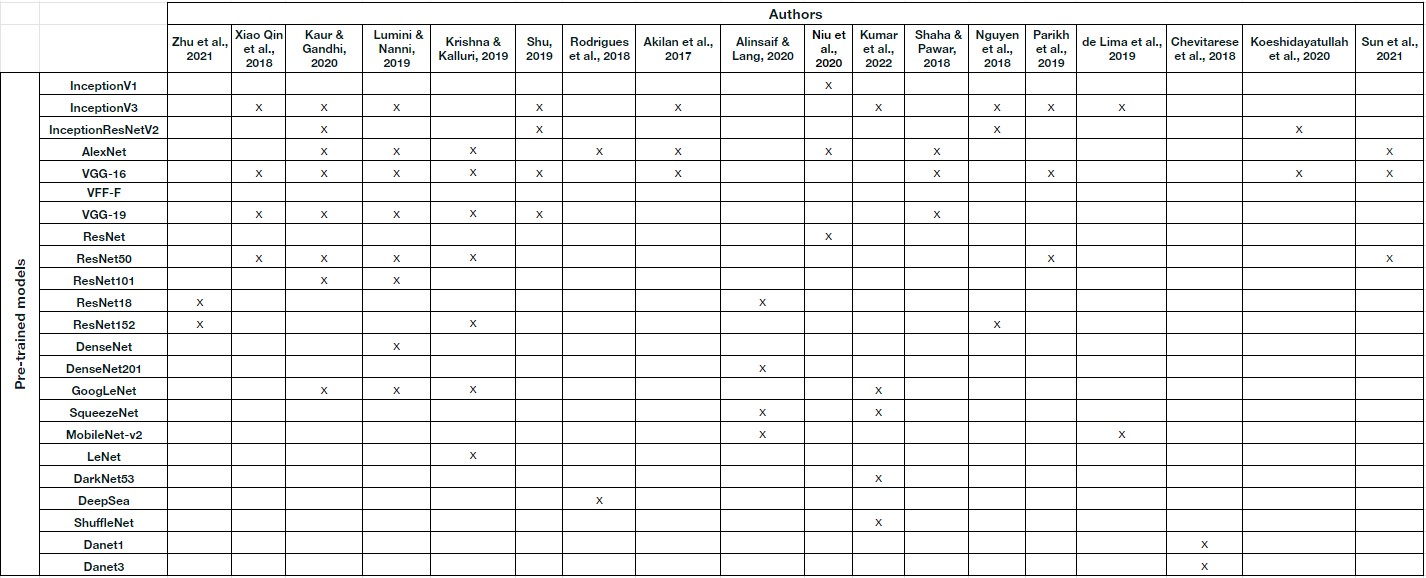}
\end{table*}

Table \ref{tab:Table3} summarizes the datasets in which different works evaluated different pre-trained models for FE in the literature reviewed in this work.

\begin{table*}[!ht]
\centering
\caption{Summary of literature review - datasets}
\label{tab:Table3}
\includegraphics[scale=.55, angle=90]{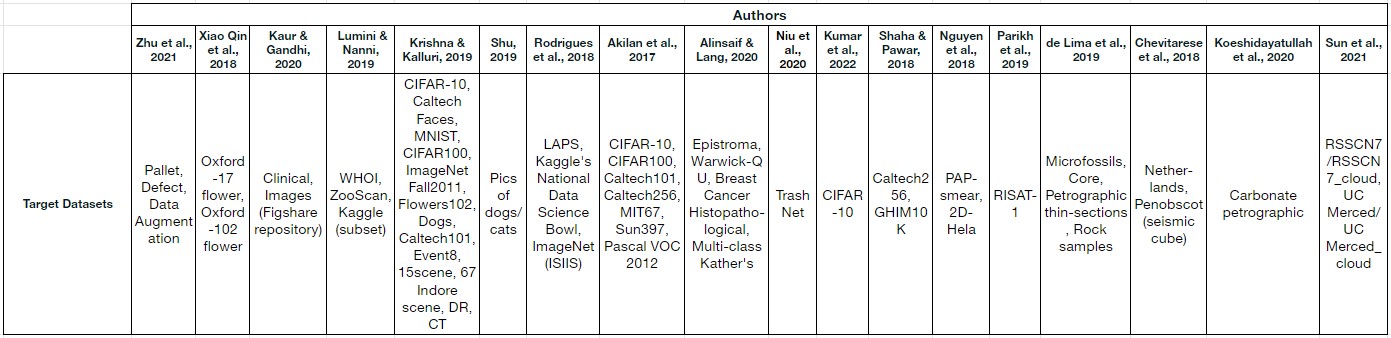}
\end{table*}

\section{Experiments} \label{sec:experiments}

In this section, we discuss the experiments to evaluate different pre-trained models in different datasets. Firstly, we present the datasets and the models used in our experiments. After, we describe the methodology adopted in these experiments. Finally, we present and discuss the results of the experiments.

\subsection{Datasets}
\label{datasets}
Image datasets are collections of digital images that can be both annotated or not. In general, image datasets can be used for various computer vision tasks, such as image recognition, object detection, segmentation, object tracking, etc. These datasets may contain images of different sizes, resolutions, and formats, and they are often annotated with additional information such as object classes, bounding boxes, segmentation masks, etc. They are essential for training and evaluating computer vision models, and their availability has enabled significant advances in the field of computer vision in recent years. These datasets are often used to benchmark new algorithms and to improve the performance of existing ones.

There are several widely used image datasets in computer vision research. In this work, the following ones were considered for image classification, since they are widely used in the literature, are colorful, and have different characteristics: Geological Images dataset \cite{TodescatoGBC23}, Stanford Cars \cite{KrauseStarkDengFei:13}, CIFAR-10 \cite{krizhevsky:09}, and STL10 \cite{coates:11}.



The main focus of this work is to evaluate the performance of the pre-trained models over the Geological Images dataset. However, as a secondary objective, Stanford Cars, CIFAR-10, and STL10 datasets were also included in our evaluation, in order to carry out a more comprehensive evaluation of the models. Table \ref{tab:datasetTable} shows the main information of all datasets used in this work. 

\begin{table*}[!ht]
\centering
\caption{Datasets Information}
\label{tab:datasetTable}
\resizebox{\textwidth}{!}{%
\begin{tabular}{|l|c|c|c|}
\hline
\textbf{Dataset}              & \textbf{Instances} & \textbf{Classes} & \textbf{Average Instances per Class $\pm$ Std} \\ \hline
\textbf{Geological Images \cite{TodescatoGBC23}}    & 25725              & 45               & 571,67 $\pm$ 1290,90                          \\ \hline
\textbf{Stanford Cars \cite{KrauseStarkDengFei:13}}        & 16185              & 196              & 84 $\pm$ 6,28                           \\ \hline
\textbf{CIFAR-10 \cite{krizhevsky:09}}             & 60000              & 10               & 6000 $\pm$ 0                                   \\ \hline
\textbf{STL10 \cite{coates:11}}               & 100000             & 10               & 10000 $\pm$ 0                                  \\ \hline
\end{tabular}%
}
\end{table*}



Regarding the selected datasets, CIFAR-10 and STL10 are balanced and include sets of images of homogeneous size. The Geological images and the Stanford Cars datasets are unbalanced (Stanford Cars is slightly unbalanced) and have images of heterogeneous sizes. In the following, we will provide more details regarding each dataset adopted in this study.


\subsubsection{Geological Images dataset} \label{geo}

\begin{figure}
\centering
\includegraphics[width=.6\textwidth]{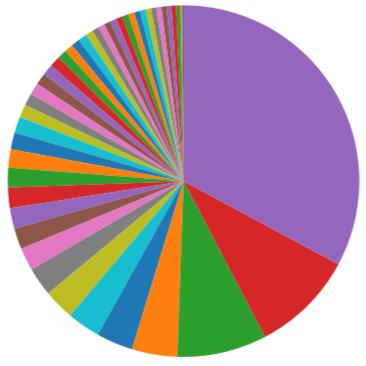}
\caption{Illustration of the imbalance in the number of images for each class on the Geological Images dataset \cite{TodescatoGBC23}.}
\label{fig:geo}
\end{figure}

\begin{figure}
\centering
\includegraphics[width=.7\textwidth]{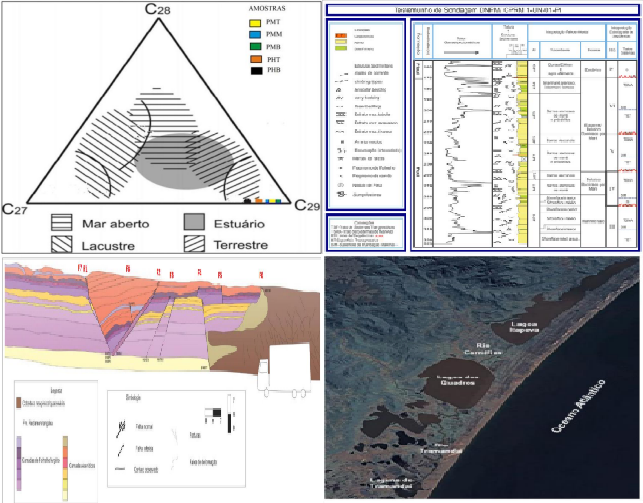}
\caption{Examples of images included in the Geological Images dataset \cite{TodescatoGBC23}.}
\label{fig:geoEx}
\end{figure}

\begin{figure}[!ht]
\centering
\includegraphics[width=.7\textwidth]{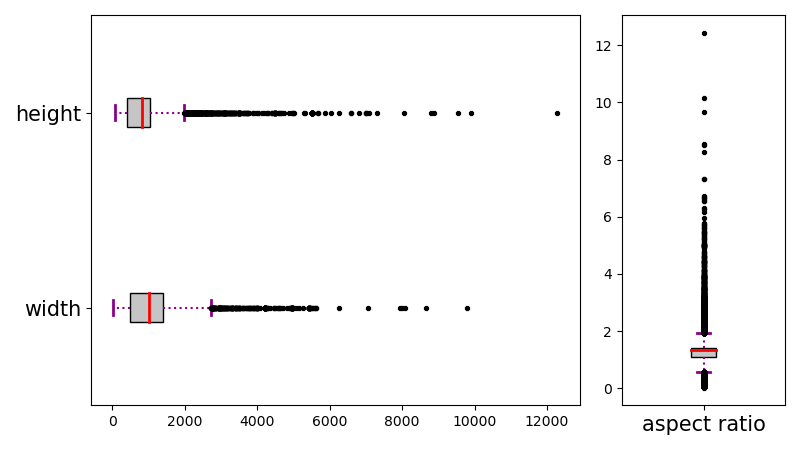}
\caption{Statistics of the distribution of height, width, and aspect ratio of images in the Geological Images dataset \cite{TodescatoGBC23}.}
\label{fig:geoRatio}
\end{figure}

This is a domain-specific dataset \cite{abel2019knowledge} that includes a set of annotated images that are relevant for applications in Geosciences. The adopted image dataset includes images extracted from public and private sources. The images were obtained from the Petrobras Geoscience Bulletins, Theses, and Dissertations from the Geosciences Institute of UFRGS, as well as from several sources on the internet. Cleaning procedures were performed to refine the set of images, aiming to remove invalid and/or non-relevant images. In total, the dataset includes 25,725 annotated images of 45 different classes. This dataset presents an imbalance in the distribution of images across classes, where there are classes with only 36 images and classes with 8450 images. The average of images per class is 571,67 with a standard deviation of 1290,90. The distribution of instances per class of this dataset is represented in Figure \ref{fig:geo}. Figure \ref{fig:geoEx} shows a set of images that are included in the Geological Images dataset. Besides that, the images included in this dataset are heterogeneous regarding height, width, and aspect ratio, as is represented in Figure \ref{fig:geoRatio}.

\subsubsection{Stanford Cars} \label{cars}

\begin{figure}[!ht]
\centering
\includegraphics[width=.7\textwidth]{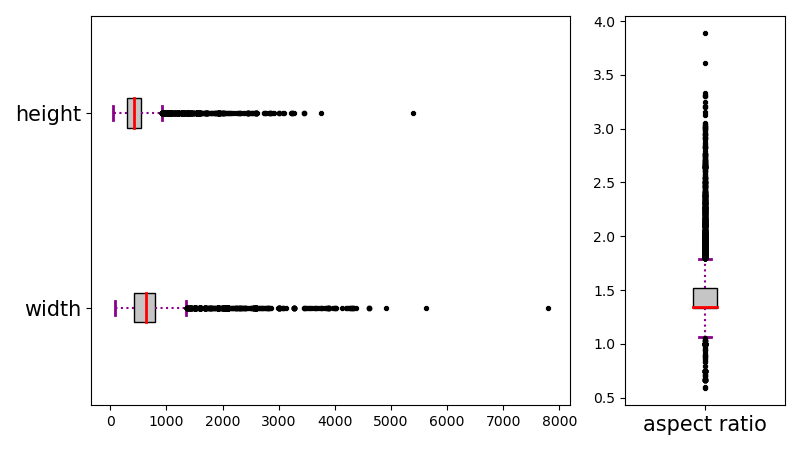}
\caption{Statistics of the distribution of height, width, and aspect ratio of images in the Stanford Cars dataset \cite{TodescatoGBC23}.}
\label{fig:stanfordRatio}
\end{figure}

\begin{figure}[!ht]
\centering
\includegraphics[width=.5\textwidth]{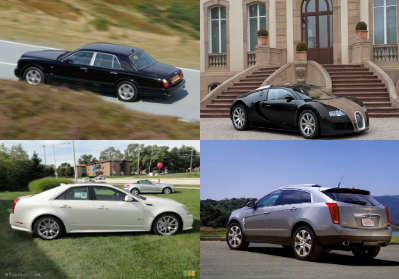}
\caption{Images included in the Stanford Cars dataset \cite{KrauseStarkDengFei:13}.}
\label{fig:stanfordEx}
\end{figure}

It consists of 16,185 images of 196 different car classes, where each car class has been carefully labeled with its make, model, and year. The images were collected from various online sources, including auction websites, car dealerships, and online classifieds.
Each image in the dataset \footnote{ \url{https://ai.stanford.edu/~jkrause/cars/car_dataset.html}} is associated with a text file that contains metadata about the car, including the car's make, model, and year, as well as its class ID, which ranges from 1 to 196. The dataset also includes a set of bounding boxes for each image, which identify the location of the car within the image. Figure \ref{fig:stanfordEx} shows a set of images that are included in the Stanford Cars dataset. Also, the images included in Stanford Cars dataset compared to Geological Images dataset are partially heterogeneous regarding height, width, and aspect ratio, as is represented in Figure \ref{fig:stanfordRatio}.

\subsubsection{CIFAR-10} \label{cifar}

\begin{figure}[!ht]
\centering
\includegraphics[width=.5\textwidth]{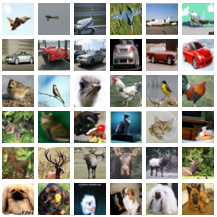}
\caption{Images included in the CIFAR-10 dataset \cite{krizhevsky:09}.}
\label{fig:cifar10Ex}
\end{figure}

The dataset \footnote{ \url{https://www.cs.toronto.edu/~kriz/cifar.html}} contains 60,000 32x32 color images in 10 different classes. Each class contains 6,000 images, and the dataset is split into 50,000 training images and 10,000 test images. The classes in the dataset are airplane, automobile, bird, cat, deer, dog, frog, horse, ship, and truck. Figure \ref{fig:cifar10Ex} shows a set of images that are included in the CIFAR-10 dataset.

\subsubsection{STL10} \label{stl10}

\begin{figure}[!ht]
\centering
\includegraphics[width=.5\textwidth]{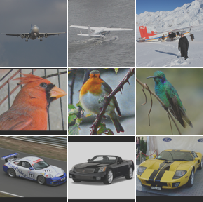}
\caption{Images included in the STL10 dataset \cite{coates:11}.}
\label{fig:stl10Ex}
\end{figure}

This is an image recognition dataset\footnote{ \url{https://cs.stanford.edu/~acoates/stl10/}} that contains a set of 10 classes, each represented by 1,000 labeled images. The images in the dataset are 96x96 pixels and are derived from a larger set of unlabeled images collected from the Internet. STL10 was developed to be a more challenging dataset for image recognition tasks, as the images are larger than those in other popular datasets like CIFAR-10 and ImageNet. The 10 classes are airplane, bird, car, cat, deer, dog, horse, monkey, ship, and truck. It is also divided into a training set of 5,000 images and a test set of 8,000 images, with each class evenly represented in both sets. In addition to the labeled images, the STL10 dataset also includes an unlabeled set of 100,000 images, which can be used for unsupervised learning tasks like FE and clustering. In this work, we adopted only the labeled images. Figure \ref{fig:stl10Ex} shows a set of images that are included in the STL10 dataset.

\subsection{Pre-trained Models}
\label{models}

Due to the increasing in adoption of transfer learning, nowadays there are several pre-trained models available in the literature. In our work, we use a wide range of pre-trained models available in repositories\footnote{Can be accessed through \url{https://pytorch.org/vision/stable/models.html} and \url{https://github.com/openai/CLIP}}.
The majority of these models considered in this work were pre-trained using ImageNet-1K\footnote{Can be accessed through \url{https://image-net.org/index.php}}\cite{deng:09} dataset except for the CLIP\cite{radford:21} based models that were pre-trained in a dataset with 400 million images called WebImageText (WIT). Table \ref{tab:modelsTable} presents the following properties of the selected models: number of output features, number of parameters, training dataset, and architecture. Notice that, clip-rn50 and clip-vit-b adopt two encoders, one for images and the other for text. Thus, in Table \ref{tab:modelsTable}, the notation CNN + Tr means that the encoder for images is based on CNN and the encoder for texts is based on transformers.

\begin{table}[!t]
\centering
\caption{Pre-trained Models Information. CNN indicates a convolutional neural networks architecture and Tr indicates a transformer-based architecture.}
\label{tab:modelsTable}

\begin{tabular}{|c|c|c|c|}
\hline
\textbf{Models} & \multicolumn{1}{l|}{\textbf{\begin{tabular}[c]{@{}l@{}}Output\\ Features\end{tabular}}} & \textbf{Parameters}  & \textbf{Architecture} \\ \hline
\textbf{alexnet\cite{krizhevsky:14}}               & 256  & 61,100,840                       & CNN                       \\ \hline
\textbf{clip\_rn50\cite{he:16,radford:21}}            & 1024 & 63,000,000   & CNN + Tr  \\ \hline
\textbf{clip\_vit\_b\cite{radford:21}}          & 512  & 63,000,000   & Tr + Tr \\ \hline
\textbf{convnext\_large\cite{liu:22}}       & 1536 & 197,767,336                      & CNN                       \\ \hline
\textbf{densenet161\cite{huang:17}}           & 2208 & 28,681,000                      & CNN                       \\ \hline
\textbf{googlenet\cite{szegedy:15}}             & 1000 & 6,624,904                       & CNN                       \\ \hline
\textbf{inception\_v3\cite{szegedy:16}}         & 1000 & 27,161,264                      & CNN                       \\ \hline
\textbf{mnasnet1\_3\cite{tan:19}}           & 1000 & 6,282,256                     & CNN                       \\ \hline
\textbf{mobilenet\_v3\_large\cite{howard:19}}  & 960  & 5,483,032                        & CNN                       \\ \hline
\textbf{regnet\_y\_3\_2gf\cite{radosavovic:20}}     & 1000 & 19,436,338                       & CNN                       \\ \hline
\textbf{resnext101\_64x4d\cite{xie:17}}     & 1000 & 83,455,272                     & CNN                       \\ \hline
\textbf{shufflenet\_v2\_x2\_0\cite{ma:18}} & 1000 & 7,393,996                       & CNN                       \\ \hline
\textbf{squeezenet1\_1\cite{iandola:16}}        & 512  & 1,235,496                       & CNN                       \\ \hline
\textbf{vgg19\_bn\cite{simonyanAndzisserman:14}}             & 512  & 143,678,248                    & CNN                       \\ \hline
\textbf{vit\_h\_14\cite{dosovitskiy:20}}            & 1000 & 632,045,800                     & Tr               \\ \hline
\textbf{wide\_resnet101\_2\cite{zagoruyko:16}}    & 1000 & 126,886,696                     & CNN                       \\ \hline
\end{tabular}%

\end{table}

\subsection{Methodology}
\label{methodo}

Our goal is to evaluate the performance of different available pre-trained models as feature extractors in the image classification task in different datasets. We used the datasets and models previously detailed for conducting our experiments. Notice also that, since there are different versions available for each family of models, we have selected a single model for each family that presented the best overall performances according to the literature.

Since we are considering 16 models and four datasets, a total of 64 experiments considering pairs of models and datasets were performed.

In each experiment, each pre-trained model was used as a feature extractor. Therefore, in this context, all initial layers (except the last one) of the model, responsible for extracting relevant features from the input images, were maintained, while the last layer was replaced by a new output layer, with $N$ units (where $N$ is proportional to the number of classes in the used dataset) and a linear activation function and a softmax layer. During the model training, the weights of the initial layers (responsible for extracting features) are not adjusted (they are kept "frozen"), while the weights of the last layer are adjusted. That is, the pre-trained model is used just for extracting features.

For each experiment, the datasets went through a homogeneous pre-processing. The pre-processing consisted of applying resizing, center cropping, and normalization. The resize is always done by decreasing or increasing the size of the smallest dimension of the image to the size of the pre-trained model's input. Then the center crop is performed, where the central area of the image is cut as a square that matches the size of the input, and this information is used for representing the whole image. Finally, we ensure that all images are converted to RGB. 

Each model was evaluated considering a 5-fold cross-validation process. We adopted the following metrics to evaluate our approach: accuracy, macro and weighted f1-score. Thus, the reported metrics are averages obtained considering the performance in each test fold of this cross-validation process.

Regarding the training hyperparameters, the \emph{learning rate} used in this study was $0.001$ with a \emph{momentum} of $0.9$. We adopted the \emph{Adam optimizer} with default parameters, with the \emph{Cross-Entropy} loss function. All executions were done using $100$ epochs and \emph{early stopping} with a minimal improvement of $0.001$ and patience of $5$.

\subsection{Results} \label{sec:results}

The following tables represent the model's performance according to the selected metrics for each dataset. Table \ref{tab:geoMetricTable} represents the model's evaluation on the Geological Images dataset. Table \ref{tab:stanfordMetricTable} represents the model's evaluation according to Stanford Cars dataset. Table \ref{tab:cifarMetricTable} represents the model's evaluation considering the CIFAR-10 dataset. Table \ref{tab:stlMetricTable} represents the model's evaluation for the STL10 dataset. In each table, we highlight the model with the best performance in green and with the less performance in red. 

\begin{table*}[!ht]
\centering
\caption{Geological Images Dataset}
\label{tab:geoMetricTable}
\resizebox{\textwidth}{!}{%
\begin{tabular}{l|c|c|c|c|c|c|c}
\hline
\multicolumn{8}{|c|}{\textbf{Geological Images Dataset}} \\ \hline
\multicolumn{1}{|l|}{} &
  \multicolumn{1}{l|}{} &
  \multicolumn{3}{c|}{\textbf{Macro}} &
  \multicolumn{3}{c|}{\textbf{Weighted}} \\ \hline
\multicolumn{1}{|l|}{\textbf{Model \textbackslash Metrics}} &
  \multicolumn{1}{c|}{\textbf{Accuracy}} &
  \multicolumn{1}{c|}{\textbf{Precision}} &
  \multicolumn{1}{c|}{\textbf{Recall}} &
  \multicolumn{1}{c|}{\textbf{F-Score}} &
  \multicolumn{1}{c|}{\textbf{Precision}} &
  \multicolumn{1}{c|}{\textbf{Recall}} &
  \multicolumn{1}{c|}{\textbf{F-Score}} \\ \hline
\multicolumn{1}{|l|}{\textbf{alexnet}} &
  \multicolumn{1}{l|}{0,85} &
  \multicolumn{1}{l|}{0,74} &
  \multicolumn{1}{l|}{0,67} &
  \multicolumn{1}{l|}{0,69} &
  \multicolumn{1}{l|}{0,84} &
  \multicolumn{1}{l|}{0,85} &
  \multicolumn{1}{l|}{0,84} \\ \hline
\multicolumn{1}{|l|}{\textbf{clip\_rn50}} &
  \multicolumn{1}{l|}{0,93} &
  \multicolumn{1}{l|}{0,86} &
  \multicolumn{1}{l|}{0,83} &
  \multicolumn{1}{l|}{0,84} &
  \multicolumn{1}{l|}{0,92} &
  \multicolumn{1}{l|}{0,93} &
  \multicolumn{1}{l|}{0,92} \\ \hline
\multicolumn{1}{|l|}{\textbf{clip\_vit\_b}} &
  \multicolumn{1}{l|}{\color[HTML]{34FF34} 0,93} &
  \multicolumn{1}{l|}{\color[HTML]{34FF34} 0,86} &
  \multicolumn{1}{l|}{\color[HTML]{34FF34} 0,83} &
  \multicolumn{1}{l|}{\color[HTML]{34FF34} 0,84} &
  \multicolumn{1}{l|}{\color[HTML]{34FF34} 0,93} &
  \multicolumn{1}{l|}{\color[HTML]{34FF34} 0,93} &
  \multicolumn{1}{l|}{\color[HTML]{34FF34} 0,93} \\ \hline
\multicolumn{1}{|l|}{\textbf{convnext\_large}} &
  \multicolumn{1}{l|}{0,91} &
  \multicolumn{1}{l|}{0,84} &
  \multicolumn{1}{l|}{0,80} &
  \multicolumn{1}{l|}{0,82} &
  \multicolumn{1}{l|}{0,91} &
  \multicolumn{1}{l|}{0,91} &
  \multicolumn{1}{l|}{0,91} \\ \hline
\multicolumn{1}{|l|}{\textbf{densenet161}} &
  \multicolumn{1}{l|}{0,90} &
  \multicolumn{1}{l|}{0,83} &
  \multicolumn{1}{l|}{0,78} &
  \multicolumn{1}{l|}{0,80} &
  \multicolumn{1}{l|}{0,90} &
  \multicolumn{1}{l|}{0,90} &
  \multicolumn{1}{l|}{0,90} \\ \hline
\multicolumn{1}{|l|}{\textbf{googlenet}} &
  \multicolumn{1}{l|}{0,87} &
  \multicolumn{1}{l|}{0,75} &
  \multicolumn{1}{l|}{0,72} &
  \multicolumn{1}{l|}{0,73} &
  \multicolumn{1}{l|}{0,86} &
  \multicolumn{1}{l|}{0,87} &
  \multicolumn{1}{l|}{0,86} \\ \hline
\multicolumn{1}{|l|}{\textbf{inception\_v3}} &
  \multicolumn{1}{l|}{\color[HTML]{FE0000} 0,83} &
  \multicolumn{1}{l|}{\color[HTML]{FE0000} 0,70} &
  \multicolumn{1}{l|}{\color[HTML]{FE0000} 0,65} &
  \multicolumn{1}{l|}{\color[HTML]{FE0000} 0,67} &
  \multicolumn{1}{l|}{\color[HTML]{FE0000} 0,82} &
  \multicolumn{1}{l|}{\color[HTML]{FE0000} 0,83} &
  \multicolumn{1}{l|}{\color[HTML]{FE0000} 0,83} \\ \hline
\multicolumn{1}{|l|}{\textbf{mnasnet1\_3}} &
  \multicolumn{1}{l|}{0,88} &
  \multicolumn{1}{l|}{0,77} &
  \multicolumn{1}{l|}{0,73} &
  \multicolumn{1}{l|}{0,75} &
  \multicolumn{1}{l|}{0,87} &
  \multicolumn{1}{l|}{0,88} &
  \multicolumn{1}{l|}{0,87} \\ \hline
\multicolumn{1}{|l|}{\textbf{mobilenet\_v3\_large}} &
  \multicolumn{1}{l|}{0,90} &
  \multicolumn{1}{l|}{0,82} &
  \multicolumn{1}{l|}{0,77} &
  \multicolumn{1}{l|}{0,79} &
  \multicolumn{1}{l|}{0,90} &
  \multicolumn{1}{l|}{0,90} &
  \multicolumn{1}{l|}{0,90} \\ \hline
\multicolumn{1}{|l|}{\textbf{regnet\_y\_3\_2gf}} &
  \multicolumn{1}{l|}{0,89} &
  \multicolumn{1}{l|}{0,79} &
  \multicolumn{1}{l|}{0,76} &
  \multicolumn{1}{l|}{0,77} &
  \multicolumn{1}{l|}{0,89} &
  \multicolumn{1}{l|}{0,89} &
  \multicolumn{1}{l|}{0,89} \\ \hline
\multicolumn{1}{|l|}{\textbf{resnext101\_64x4d}} &
  \multicolumn{1}{l|}{0,88} &
  \multicolumn{1}{l|}{0,79} &
  \multicolumn{1}{l|}{0,74} &
  \multicolumn{1}{l|}{0,76} &
  \multicolumn{1}{l|}{0,88} &
  \multicolumn{1}{l|}{0,88} &
  \multicolumn{1}{l|}{0,88} \\ \hline
\multicolumn{1}{|l|}{\textbf{shufflenet\_v2\_x2\_0}} &
  \multicolumn{1}{l|}{0,89} &
  \multicolumn{1}{l|}{0,80} &
  \multicolumn{1}{l|}{0,76} &
  \multicolumn{1}{l|}{0,78} &
  \multicolumn{1}{l|}{0,89} &
  \multicolumn{1}{l|}{0,89} &
  \multicolumn{1}{l|}{0,89} \\ \hline
\multicolumn{1}{|l|}{\textbf{squeezenet1\_1}} &
  \multicolumn{1}{l|}{0,87} &
  \multicolumn{1}{l|}{0,77} &
  \multicolumn{1}{l|}{0,72} &
  \multicolumn{1}{l|}{0,74} &
  \multicolumn{1}{l|}{0,87} &
  \multicolumn{1}{l|}{0,87} &
  \multicolumn{1}{l|}{0,87} \\ \hline
\multicolumn{1}{|l|}{\textbf{vgg19\_bn}} &
  \multicolumn{1}{l|}{0,88} &
  \multicolumn{1}{l|}{0,79} &
  \multicolumn{1}{l|}{0,74} &
  \multicolumn{1}{l|}{0,76} &
  \multicolumn{1}{l|}{0,88} &
  \multicolumn{1}{l|}{0,88} &
  \multicolumn{1}{l|}{0,88} \\ \hline
\multicolumn{1}{|l|}{\textbf{vit\_h\_14}} &
  \multicolumn{1}{l|}{0,91} &
  \multicolumn{1}{l|}{0,82} &
  \multicolumn{1}{l|}{0,79} &
  \multicolumn{1}{l|}{0,80} &
  \multicolumn{1}{l|}{0,90} &
  \multicolumn{1}{l|}{0,91} &
  \multicolumn{1}{l|}{0,90} \\ \hline
\multicolumn{1}{|l|}{\textbf{wide\_resnet101\_2}} &
  \multicolumn{1}{l|}{0,89} &
  \multicolumn{1}{l|}{0,79} &
  \multicolumn{1}{l|}{0,75} &
  \multicolumn{1}{l|}{0,77} &
  \multicolumn{1}{l|}{0,89} &
  \multicolumn{1}{l|}{0,89} &
  \multicolumn{1}{l|}{0,89} \\ \hline
 &
   &
   &
   &
   &
   &
   &
   \\ \hline
\multicolumn{1}{|l|}{\textbf{Average}} &
  \multicolumn{1}{l|}{0,89} &
  \multicolumn{1}{l|}{0,79} &
  \multicolumn{1}{l|}{0,75} &
  \multicolumn{1}{l|}{0,77} &
  \multicolumn{1}{l|}{0,88} &
  \multicolumn{1}{l|}{0,89} &
  \multicolumn{1}{l|}{0,88} \\ \hline
\multicolumn{1}{|l|}{\textbf{Standard Deviation}} &
  \multicolumn{1}{l|}{0,02} &
  \multicolumn{1}{l|}{0,04} &
  \multicolumn{1}{l|}{0,05} &
  \multicolumn{1}{l|}{0,05} &
  \multicolumn{1}{l|}{0,03} &
  \multicolumn{1}{l|}{0,02} &
  \multicolumn{1}{l|}{0,03} \\ \hline
\end{tabular}%
}
\end{table*}

\begin{table*}[!ht]
\centering
\caption{Stanford Cars Dataset}
\label{tab:stanfordMetricTable}
\resizebox{\textwidth}{!}{%
\begin{tabular}{l|c|c|c|c|c|c|c}
\hline
\multicolumn{8}{|c|}{\textbf{Stanford Cars Dataset}} \\ \hline
\multicolumn{1}{|l|}{} &
  \multicolumn{1}{l|}{} &
  \multicolumn{3}{c|}{\textbf{Macro}} &
  \multicolumn{3}{c|}{\textbf{Weighted}} \\ \hline
\multicolumn{1}{|l|}{\textbf{Model \textbackslash Metrics}} &
  \multicolumn{1}{c|}{\textbf{Accuracy}} &
  \multicolumn{1}{c|}{\textbf{Precision}} &
  \multicolumn{1}{c|}{\textbf{Recall}} &
  \multicolumn{1}{c|}{\textbf{F-Score}} &
  \multicolumn{1}{c|}{\textbf{Precision}} &
  \multicolumn{1}{c|}{\textbf{Recall}} &
  \multicolumn{1}{c|}{F-Score} \\ \hline
\multicolumn{1}{|l|}{\textbf{alexnet}} &
  \multicolumn{1}{l|}{\color[HTML]{FE0000} 0,28} &
  \multicolumn{1}{l|}{\color[HTML]{FE0000} 0,26} &
  \multicolumn{1}{l|}{\color[HTML]{FE0000} 0,28} &
  \multicolumn{1}{l|}{\color[HTML]{FE0000} 0,26} &
  \multicolumn{1}{l|}{\color[HTML]{FE0000} 0,26} &
  \multicolumn{1}{l|}{\color[HTML]{FE0000} 0,28} &
  \multicolumn{1}{l|}{\color[HTML]{FE0000} 0,26} \\ \hline
\multicolumn{1}{|l|}{\textbf{clip\_rn50}} &
  \multicolumn{1}{l|}{0,82} &
  \multicolumn{1}{l|}{0,82} &
  \multicolumn{1}{l|}{0,82} &
  \multicolumn{1}{l|}{0,82} &
  \multicolumn{1}{l|}{0,82} &
  \multicolumn{1}{l|}{0,82} &
  \multicolumn{1}{l|}{0,82} \\ \hline
\multicolumn{1}{|l|}{\textbf{clip\_vit\_b}} &
  \multicolumn{1}{l|}{0,83} &
  \multicolumn{1}{l|}{0,83} &
  \multicolumn{1}{l|}{0,83} &
  \multicolumn{1}{l|}{0,83} &
  \multicolumn{1}{l|}{0,83} &
  \multicolumn{1}{l|}{0,83} &
  \multicolumn{1}{l|}{0,83} \\ \hline
\multicolumn{1}{|l|}{\textbf{convnext\_large}} &
  \multicolumn{1}{l|}{0,65} &
  \multicolumn{1}{l|}{0,65} &
  \multicolumn{1}{l|}{0,64} &
  \multicolumn{1}{l|}{0,64} &
  \multicolumn{1}{l|}{0,65} &
  \multicolumn{1}{l|}{0,65} &
  \multicolumn{1}{l|}{0,64} \\ \hline
\multicolumn{1}{|l|}{\textbf{densenet161}} &
  \multicolumn{1}{l|}{0,64} &
  \multicolumn{1}{l|}{0,64} &
  \multicolumn{1}{l|}{0,64} &
  \multicolumn{1}{l|}{0,64} &
  \multicolumn{1}{l|}{0,64} &
  \multicolumn{1}{l|}{0,64} &
  \multicolumn{1}{l|}{0,64} \\ \hline
\multicolumn{1}{|l|}{\textbf{googlenet}} &
  \multicolumn{1}{l|}{0,41} &
  \multicolumn{1}{l|}{0,41} &
  \multicolumn{1}{l|}{0,41} &
  \multicolumn{1}{l|}{0,41} &
  \multicolumn{1}{l|}{0,40} &
  \multicolumn{1}{l|}{0,41} &
  \multicolumn{1}{l|}{0,40} \\ \hline
\multicolumn{1}{|l|}{\textbf{inception\_v3}} &
  \multicolumn{1}{l|}{0,34} &
  \multicolumn{1}{l|}{0,33} &
  \multicolumn{1}{l|}{0,34} &
  \multicolumn{1}{l|}{0,33} &
  \multicolumn{1}{l|}{0,33} &
  \multicolumn{1}{l|}{0,34} &
  \multicolumn{1}{l|}{0,33} \\ \hline
\multicolumn{1}{|l|}{\textbf{mnasnet1\_3}} &
  \multicolumn{1}{l|}{0,42} &
  \multicolumn{1}{l|}{0,42} &
  \multicolumn{1}{l|}{0,42} &
  \multicolumn{1}{l|}{0,42} &
  \multicolumn{1}{l|}{0,41} &
  \multicolumn{1}{l|}{0,42} &
  \multicolumn{1}{l|}{0,41} \\ \hline
\multicolumn{1}{|l|}{\textbf{mobilenet\_v3\_large}} &
  \multicolumn{1}{l|}{0,56} &
  \multicolumn{1}{l|}{0,56} &
  \multicolumn{1}{l|}{0,56} &
  \multicolumn{1}{l|}{0,55} &
  \multicolumn{1}{l|}{0,56} &
  \multicolumn{1}{l|}{0,56} &
  \multicolumn{1}{l|}{0,55} \\ \hline
\multicolumn{1}{|l|}{\textbf{regnet\_y\_3\_2gf}} &
  \multicolumn{1}{l|}{0,49} &
  \multicolumn{1}{l|}{0,49} &
  \multicolumn{1}{l|}{0,49} &
  \multicolumn{1}{l|}{0,49} &
  \multicolumn{1}{l|}{0,49} &
  \multicolumn{1}{l|}{0,49} &
  \multicolumn{1}{l|}{0,49} \\ \hline
\multicolumn{1}{|l|}{\textbf{resnext101\_64x4d}} &
  \multicolumn{1}{l|}{0,35} &
  \multicolumn{1}{l|}{0,35} &
  \multicolumn{1}{l|}{0,35} &
  \multicolumn{1}{l|}{0,34} &
  \multicolumn{1}{l|}{0,34} &
  \multicolumn{1}{l|}{0,35} &
  \multicolumn{1}{l|}{0,34} \\ \hline
\multicolumn{1}{|l|}{\textbf{shufflenet\_v2\_x2\_0}} &
  \multicolumn{1}{l|}{0,50} &
  \multicolumn{1}{l|}{0,50} &
  \multicolumn{1}{l|}{0,50} &
  \multicolumn{1}{l|}{0,50} &
  \multicolumn{1}{l|}{0,50} &
  \multicolumn{1}{l|}{0,50} &
  \multicolumn{1}{l|}{0,50} \\ \hline
\multicolumn{1}{|l|}{\textbf{squeezenet1\_1}} &
  \multicolumn{1}{l|}{0,42} &
  \multicolumn{1}{l|}{0,42} &
  \multicolumn{1}{l|}{0,42} &
  \multicolumn{1}{l|}{0,41} &
  \multicolumn{1}{l|}{0,41} &
  \multicolumn{1}{l|}{0,42} &
  \multicolumn{1}{l|}{0,41} \\ \hline
\multicolumn{1}{|l|}{\textbf{vgg19\_bn}} &
  \multicolumn{1}{l|}{0,51} &
  \multicolumn{1}{l|}{0,50} &
  \multicolumn{1}{l|}{0,51} &
  \multicolumn{1}{l|}{0,50} &
  \multicolumn{1}{l|}{0,50} &
  \multicolumn{1}{l|}{0,51} &
  \multicolumn{1}{l|}{0,50} \\ \hline
\multicolumn{1}{|l|}{\textbf{vit\_h\_14}} &
  \multicolumn{1}{l|}{\color[HTML]{34FF34} 0,86} &
  \multicolumn{1}{l|}{\color[HTML]{34FF34} 0,86} &
  \multicolumn{1}{l|}{\color[HTML]{34FF34} 0,85} &
  \multicolumn{1}{l|}{\color[HTML]{34FF34} 0,85} &
  \multicolumn{1}{l|}{\color[HTML]{34FF34} 0,86} &
  \multicolumn{1}{l|}{\color[HTML]{34FF34} 0,86} &
  \multicolumn{1}{l|}{\color[HTML]{34FF34} 0,86} \\ \hline
\multicolumn{1}{|l|}{\textbf{wide\_resnet101\_2}} &
  \multicolumn{1}{l|}{0,44} &
  \multicolumn{1}{l|}{0,44} &
  \multicolumn{1}{l|}{0,44} &
  \multicolumn{1}{l|}{0,44} &
  \multicolumn{1}{l|}{0,44} &
  \multicolumn{1}{l|}{0,44} &
  \multicolumn{1}{l|}{0,44} \\ \hline
 &
   &
   &
   &
   &
   &
   &
   \\ \hline
\multicolumn{1}{|l|}{\textbf{Average}} &
  \multicolumn{1}{l|}{0,53} &
  \multicolumn{1}{l|}{0,53} &
  \multicolumn{1}{l|}{0,53} &
  \multicolumn{1}{l|}{0,53} &
  \multicolumn{1}{l|}{0,53} &
  \multicolumn{1}{l|}{0,53} &
  \multicolumn{1}{l|}{0,53} \\ \hline
\multicolumn{1}{|l|}{\textbf{Standard Deviation}} &
  \multicolumn{1}{l|}{0,18} &
  \multicolumn{1}{l|}{0,18} &
  \multicolumn{1}{l|}{0,18} &
  \multicolumn{1}{l|}{0,18} &
  \multicolumn{1}{l|}{0,18} &
  \multicolumn{1}{l|}{0,18} &
  \multicolumn{1}{l|}{0,18} \\ \hline
\end{tabular}%
}
\end{table*}

\begin{table*}[!ht]
\centering
\caption{CIFAR-10 Dataset}
\label{tab:cifarMetricTable}
\resizebox{\textwidth}{!}{%
\begin{tabular}{l|c|c|c|c|c|c|c}
\hline
\multicolumn{8}{|c|}{\textbf{CIFAR-10 Dataset}} \\ \hline
\multicolumn{1}{|l|}{} &
  \multicolumn{1}{l|}{} &
  \multicolumn{3}{c|}{\textbf{Macro}} &
  \multicolumn{3}{c|}{\textbf{Weighted}} \\ \hline
\multicolumn{1}{|l|}{\textbf{Model \textbackslash Metrics}} &
  \multicolumn{1}{c|}{\textbf{Accuracy}} &
  \multicolumn{1}{c|}{\textbf{Precision}} &
  \multicolumn{1}{c|}{\textbf{Recall}} &
  \multicolumn{1}{c|}{\textbf{F-Score}} &
  \multicolumn{1}{c|}{\textbf{Precision}} &
  \multicolumn{1}{c|}{\textbf{Recall}} &
  \multicolumn{1}{c|}{\textbf{F-Score}} \\ \hline
\multicolumn{1}{|l|}{\textbf{alexnet}} &
  \multicolumn{1}{c|}{\color[HTML]{FE0000} 0,79} &
  \multicolumn{1}{c|}{\color[HTML]{FE0000} 0,79} &
  \multicolumn{1}{c|}{\color[HTML]{FE0000} 0,79} &
  \multicolumn{1}{c|}{\color[HTML]{FE0000} 0,79} &
  \multicolumn{1}{c|}{\color[HTML]{FE0000} 0,79} &
  \multicolumn{1}{c|}{\color[HTML]{FE0000} 0,79} &
  \multicolumn{1}{c|}{\color[HTML]{FE0000} 0,79} \\ \hline
\multicolumn{1}{|l|}{\textbf{clip\_rn50}} &
  \multicolumn{1}{c|}{0,88} &
  \multicolumn{1}{c|}{0,88} &
  \multicolumn{1}{c|}{0,88} &
  \multicolumn{1}{c|}{0,88} &
  \multicolumn{1}{c|}{0,88} &
  \multicolumn{1}{c|}{0,88} &
  \multicolumn{1}{c|}{0,88} \\ \hline
\multicolumn{1}{|l|}{\textbf{clip\_vit\_b}} &
  \multicolumn{1}{c|}{0,95} &
  \multicolumn{1}{c|}{0,95} &
  \multicolumn{1}{c|}{0,95} &
  \multicolumn{1}{c|}{0,95} &
  \multicolumn{1}{c|}{0,95} &
  \multicolumn{1}{c|}{0,95} &
  \multicolumn{1}{c|}{0,95} \\ \hline
\multicolumn{1}{|l|}{\textbf{convnext\_large}} &
  \multicolumn{1}{c|}{0,96} &
  \multicolumn{1}{c|}{0,96} &
  \multicolumn{1}{c|}{0,96} &
  \multicolumn{1}{c|}{0,96} &
  \multicolumn{1}{c|}{0,96} &
  \multicolumn{1}{c|}{0,96} &
  \multicolumn{1}{c|}{0,96} \\ \hline
\multicolumn{1}{|l|}{\textbf{densenet161}} &
  \multicolumn{1}{c|}{0,93} &
  \multicolumn{1}{c|}{0,93} &
  \multicolumn{1}{c|}{0,93} &
  \multicolumn{1}{c|}{0,93} &
  \multicolumn{1}{c|}{0,93} &
  \multicolumn{1}{c|}{0,93} &
  \multicolumn{1}{c|}{0,93} \\ \hline
\multicolumn{1}{|l|}{\textbf{googlenet}} &
  \multicolumn{1}{c|}{0,87} &
  \multicolumn{1}{c|}{0,87} &
  \multicolumn{1}{c|}{0,87} &
  \multicolumn{1}{c|}{0,87} &
  \multicolumn{1}{c|}{0,87} &
  \multicolumn{1}{c|}{0,87} &
  \multicolumn{1}{c|}{0,87} \\ \hline
\multicolumn{1}{|l|}{\textbf{inception\_v3}} &
  \multicolumn{1}{c|}{0,86} &
  \multicolumn{1}{c|}{0,86} &
  \multicolumn{1}{c|}{0,86} &
  \multicolumn{1}{c|}{0,86} &
  \multicolumn{1}{c|}{0,86} &
  \multicolumn{1}{c|}{0,86} &
  \multicolumn{1}{c|}{0,86} \\ \hline
\multicolumn{1}{|l|}{\textbf{mnasnet1\_3}} &
  \multicolumn{1}{c|}{0,90} &
  \multicolumn{1}{c|}{0,90} &
  \multicolumn{1}{c|}{0,90} &
  \multicolumn{1}{c|}{0,90} &
  \multicolumn{1}{c|}{0,90} &
  \multicolumn{1}{c|}{0,90} &
  \multicolumn{1}{c|}{0,90} \\ \hline
\multicolumn{1}{|l|}{\textbf{mobilenet\_v3\_large}} &
  \multicolumn{1}{c|}{0,91} &
  \multicolumn{1}{c|}{0,91} &
  \multicolumn{1}{c|}{0,91} &
  \multicolumn{1}{c|}{0,91} &
  \multicolumn{1}{c|}{0,91} &
  \multicolumn{1}{c|}{0,91} &
  \multicolumn{1}{c|}{0,91} \\ \hline
\multicolumn{1}{|l|}{\textbf{regnet\_y\_3\_2gf}} &
  \multicolumn{1}{c|}{0,93} &
  \multicolumn{1}{c|}{0,93} &
  \multicolumn{1}{c|}{0,93} &
  \multicolumn{1}{c|}{0,93} &
  \multicolumn{1}{c|}{0,93} &
  \multicolumn{1}{c|}{0,93} &
  \multicolumn{1}{c|}{0,93} \\ \hline
\multicolumn{1}{|l|}{\textbf{resnext101\_64x4d}} &
  \multicolumn{1}{c|}{0,95} &
  \multicolumn{1}{c|}{0,95} &
  \multicolumn{1}{c|}{0,95} &
  \multicolumn{1}{c|}{0,95} &
  \multicolumn{1}{c|}{0,95} &
  \multicolumn{1}{c|}{0,95} &
  \multicolumn{1}{c|}{0,95} \\ \hline
\multicolumn{1}{|l|}{\textbf{shufflenet\_v2\_x2\_0}} &
  \multicolumn{1}{c|}{0,92} &
  \multicolumn{1}{c|}{0,92} &
  \multicolumn{1}{c|}{0,92} &
  \multicolumn{1}{c|}{0,92} &
  \multicolumn{1}{c|}{0,92} &
  \multicolumn{1}{c|}{0,92} &
  \multicolumn{1}{c|}{0,92} \\ \hline
\multicolumn{1}{|l|}{\textbf{squeezenet1\_1}} &
  \multicolumn{1}{c|}{0,85} &
  \multicolumn{1}{c|}{0,85} &
  \multicolumn{1}{c|}{0,85} &
  \multicolumn{1}{c|}{0,85} &
  \multicolumn{1}{c|}{0,85} &
  \multicolumn{1}{c|}{0,85} &
  \multicolumn{1}{c|}{0,85} \\ \hline
\multicolumn{1}{|l|}{\textbf{vgg19\_bn}} &
  \multicolumn{1}{c|}{0,88} &
  \multicolumn{1}{c|}{0,88} &
  \multicolumn{1}{c|}{0,88} &
  \multicolumn{1}{c|}{0,88} &
  \multicolumn{1}{c|}{0,88} &
  \multicolumn{1}{c|}{0,88} &
  \multicolumn{1}{c|}{0,88} \\ \hline
\multicolumn{1}{|l|}{\textbf{vit\_h\_14}} &
  \multicolumn{1}{c|}{\color[HTML]{34FF34} 0,98} &
  \multicolumn{1}{c|}{\color[HTML]{34FF34} 0,98} &
  \multicolumn{1}{c|}{\color[HTML]{34FF34} 0,98} &
  \multicolumn{1}{c|}{\color[HTML]{34FF34} 0,98} &
  \multicolumn{1}{c|}{\color[HTML]{34FF34} 0,98} &
  \multicolumn{1}{c|}{\color[HTML]{34FF34} 0,98} &
  \multicolumn{1}{c|}{\color[HTML]{34FF34} 0,98} \\ \hline
\multicolumn{1}{|l|}{\textbf{wide\_resnet101\_2}} &
  \multicolumn{1}{c|}{0,95} &
  \multicolumn{1}{c|}{0,95} &
  \multicolumn{1}{c|}{0,95} &
  \multicolumn{1}{c|}{0,95} &
  \multicolumn{1}{c|}{0,95} &
  \multicolumn{1}{c|}{0,95} &
  \multicolumn{1}{c|}{0,95} \\ \hline
 &
  \multicolumn{1}{l}{} &
  \multicolumn{1}{l}{} &
  \multicolumn{1}{l}{} &
  \multicolumn{1}{l}{} &
  \multicolumn{1}{l}{} &
  \multicolumn{1}{l}{} &
  \multicolumn{1}{l}{} \\ \hline
\multicolumn{1}{|l|}{\textbf{Average}} &
  \multicolumn{1}{l|}{0,91} &
  \multicolumn{1}{l|}{0,91} &
  \multicolumn{1}{l|}{0,91} &
  \multicolumn{1}{l|}{0,91} &
  \multicolumn{1}{l|}{0,91} &
  \multicolumn{1}{l|}{0,91} &
  \multicolumn{1}{l|}{0,91} \\ \hline
\multicolumn{1}{|l|}{\textbf{Standard Deviation}} &
  \multicolumn{1}{l|}{0,05} &
  \multicolumn{1}{l|}{0,05} &
  \multicolumn{1}{l|}{0,05} &
  \multicolumn{1}{l|}{0,05} &
  \multicolumn{1}{l|}{0,05} &
  \multicolumn{1}{l|}{0,05} &
  \multicolumn{1}{l|}{0,05} \\ \hline
\end{tabular}%
}
\end{table*}

\begin{table*}[!ht]
\centering
\caption{STL10 Dataset}
\label{tab:stlMetricTable}
\resizebox{\textwidth}{!}{%
\begin{tabular}{l|c|c|c|c|c|c|c}
\hline
\multicolumn{8}{|c|}{\textbf{STL10 Dataset}} \\ \hline
\multicolumn{1}{|l|}{} &
  \multicolumn{1}{l|}{} &
  \multicolumn{3}{c|}{\textbf{Macro}} &
  \multicolumn{3}{c|}{\textbf{Weighted}} \\ \hline
\multicolumn{1}{|l|}{\textbf{Model \textbackslash Metrics}} &
  \multicolumn{1}{c|}{\textbf{Accuracy}} &
  \multicolumn{1}{c|}{\textbf{Precision}} &
  \multicolumn{1}{c|}{\textbf{Recall}} &
  \multicolumn{1}{c|}{\textbf{F-Score}} &
  \multicolumn{1}{c|}{\textbf{Precision}} &
  \multicolumn{1}{c|}{\textbf{Recall}} &
  \multicolumn{1}{c|}{\textbf{F-Score}} \\ \hline
\multicolumn{1}{|l|}{\textbf{alexnet}} &
  \multicolumn{1}{l|}{\color[HTML]{FE0000} 0,88} &
  \multicolumn{1}{l|}{\color[HTML]{FE0000} 0,88} &
  \multicolumn{1}{l|}{\color[HTML]{FE0000} 0,88} &
  \multicolumn{1}{l|}{\color[HTML]{FE0000} 0,88} &
  \multicolumn{1}{l|}{\color[HTML]{FE0000} 0,88} &
  \multicolumn{1}{l|}{\color[HTML]{FE0000} 0,88} &
  \multicolumn{1}{l|}{\color[HTML]{FE0000} 0,88} \\ \hline
\multicolumn{1}{|l|}{\textbf{clip\_rn50}} &
  \multicolumn{1}{l|}{0,97} &
  \multicolumn{1}{l|}{0,97} &
  \multicolumn{1}{l|}{0,97} &
  \multicolumn{1}{l|}{0,97} &
  \multicolumn{1}{l|}{0,97} &
  \multicolumn{1}{l|}{0,97} &
  \multicolumn{1}{l|}{0,97} \\ \hline
\multicolumn{1}{|l|}{\textbf{clip\_vit\_b}} &
  \multicolumn{1}{l|}{0,99} &
  \multicolumn{1}{l|}{0,99} &
  \multicolumn{1}{l|}{0,99} &
  \multicolumn{1}{l|}{0,99} &
  \multicolumn{1}{l|}{0,99} &
  \multicolumn{1}{l|}{0,99} &
  \multicolumn{1}{l|}{0,99} \\ \hline
\multicolumn{1}{|l|}{\textbf{convnext\_large}} &
  \multicolumn{1}{l|}{0,99} &
  \multicolumn{1}{l|}{0,99} &
  \multicolumn{1}{l|}{0,99} &
  \multicolumn{1}{l|}{0,99} &
  \multicolumn{1}{l|}{0,99} &
  \multicolumn{1}{l|}{0,99} &
  \multicolumn{1}{l|}{0,99} \\ \hline
\multicolumn{1}{|l|}{\textbf{densenet161}} &
  \multicolumn{1}{l|}{0,98} &
  \multicolumn{1}{l|}{0,98} &
  \multicolumn{1}{l|}{0,98} &
  \multicolumn{1}{l|}{0,98} &
  \multicolumn{1}{l|}{0,98} &
  \multicolumn{1}{l|}{0,98} &
  \multicolumn{1}{l|}{0,98} \\ \hline
\multicolumn{1}{|l|}{\textbf{googlenet}} &
  \multicolumn{1}{l|}{0,96} &
  \multicolumn{1}{l|}{0,96} &
  \multicolumn{1}{l|}{0,96} &
  \multicolumn{1}{l|}{0,96} &
  \multicolumn{1}{l|}{0,96} &
  \multicolumn{1}{l|}{0,96} &
  \multicolumn{1}{l|}{0,96} \\ \hline
\multicolumn{1}{|l|}{\textbf{inception\_v3}} &
  \multicolumn{1}{l|}{0,96} &
  \multicolumn{1}{l|}{0,96} &
  \multicolumn{1}{l|}{0,96} &
  \multicolumn{1}{l|}{0,96} &
  \multicolumn{1}{l|}{0,96} &
  \multicolumn{1}{l|}{0,96} &
  \multicolumn{1}{l|}{0,96} \\ \hline
\multicolumn{1}{|l|}{\textbf{mnasnet1\_3}} &
  \multicolumn{1}{l|}{0,97} &
  \multicolumn{1}{l|}{0,97} &
  \multicolumn{1}{l|}{0,97} &
  \multicolumn{1}{l|}{0,97} &
  \multicolumn{1}{l|}{0,97} &
  \multicolumn{1}{l|}{0,97} &
  \multicolumn{1}{l|}{0,97} \\ \hline
\multicolumn{1}{|l|}{\textbf{mobilenet\_v3\_large}} &
  \multicolumn{1}{l|}{0,96} &
  \multicolumn{1}{l|}{0,96} &
  \multicolumn{1}{l|}{0,96} &
  \multicolumn{1}{l|}{0,96} &
  \multicolumn{1}{l|}{0,96} &
  \multicolumn{1}{l|}{0,96} &
  \multicolumn{1}{l|}{0,96} \\ \hline
\multicolumn{1}{|l|}{\textbf{regnet\_y\_3\_2gf}} &
  \multicolumn{1}{l|}{0,98} &
  \multicolumn{1}{l|}{0,98} &
  \multicolumn{1}{l|}{0,98} &
  \multicolumn{1}{l|}{0,98} &
  \multicolumn{1}{l|}{0,98} &
  \multicolumn{1}{l|}{0,98} &
  \multicolumn{1}{l|}{0,98} \\ \hline
\multicolumn{1}{|l|}{\textbf{resnext101\_64x4d}} &
  \multicolumn{1}{l|}{0,99} &
  \multicolumn{1}{l|}{0,99} &
  \multicolumn{1}{l|}{0,99} &
  \multicolumn{1}{l|}{0,99} &
  \multicolumn{1}{l|}{0,99} &
  \multicolumn{1}{l|}{0,99} &
  \multicolumn{1}{l|}{0,99} \\ \hline
\multicolumn{1}{|l|}{\textbf{shufflenet\_v2\_x2\_0}} &
  \multicolumn{1}{l|}{0,97} &
  \multicolumn{1}{l|}{0,97} &
  \multicolumn{1}{l|}{0,97} &
  \multicolumn{1}{l|}{0,97} &
  \multicolumn{1}{l|}{0,97} &
  \multicolumn{1}{l|}{0,97} &
  \multicolumn{1}{l|}{0,97} \\ \hline
\multicolumn{1}{|l|}{\textbf{squeezenet1\_1}} &
  \multicolumn{1}{l|}{0,91} &
  \multicolumn{1}{l|}{0,91} &
  \multicolumn{1}{l|}{0,91} &
  \multicolumn{1}{l|}{0,91} &
  \multicolumn{1}{l|}{0,91} &
  \multicolumn{1}{l|}{0,91} &
  \multicolumn{1}{l|}{0,91} \\ \hline
\multicolumn{1}{|l|}{\textbf{vgg19\_bn}} &
  \multicolumn{1}{l|}{0,96} &
  \multicolumn{1}{l|}{0,96} &
  \multicolumn{1}{l|}{0,96} &
  \multicolumn{1}{l|}{0,96} &
  \multicolumn{1}{l|}{0,96} &
  \multicolumn{1}{l|}{0,96} &
  \multicolumn{1}{l|}{0,96} \\ \hline
\multicolumn{1}{|l|}{\textbf{vit\_h\_14}} &
  \multicolumn{1}{l|}{\color[HTML]{34FF34} 1,00} &
  \multicolumn{1}{l|}{\color[HTML]{34FF34} 1,00} &
  \multicolumn{1}{l|}{\color[HTML]{34FF34} 1,00} &
  \multicolumn{1}{l|}{\color[HTML]{34FF34} 1,00} &
  \multicolumn{1}{l|}{\color[HTML]{34FF34} 1,00} &
  \multicolumn{1}{l|}{\color[HTML]{34FF34} 1,00} &
  \multicolumn{1}{l|}{\color[HTML]{34FF34} 1,00} \\ \hline
\multicolumn{1}{|l|}{\textbf{wide\_resnet101\_2}} &
  \multicolumn{1}{l|}{0,99} &
  \multicolumn{1}{l|}{0,99} &
  \multicolumn{1}{l|}{0,99} &
  \multicolumn{1}{l|}{0,99} &
  \multicolumn{1}{l|}{0,99} &
  \multicolumn{1}{l|}{0,99} &
  \multicolumn{1}{l|}{0,99} \\ \hline
 &
   &
   &
   &
   &
   &
   &
   \\ \hline
\multicolumn{1}{|l|}{\textbf{Average}} &
  \multicolumn{1}{l|}{0,97} &
  \multicolumn{1}{l|}{0,97} &
  \multicolumn{1}{l|}{0,97} &
  \multicolumn{1}{l|}{0,97} &
  \multicolumn{1}{l|}{0,97} &
  \multicolumn{1}{l|}{0,97} &
  \multicolumn{1}{l|}{0,97} \\ \hline
\multicolumn{1}{|l|}{\textbf{Standard Deviation}} &
  \multicolumn{1}{l|}{0,03} &
  \multicolumn{1}{l|}{0,03} &
  \multicolumn{1}{l|}{0,03} &
  \multicolumn{1}{l|}{0,03} &
  \multicolumn{1}{l|}{0,03} &
  \multicolumn{1}{l|}{0,03} &
  \multicolumn{1}{l|}{0,03} \\ \hline
\end{tabular}%
}
\end{table*}

\begin{figure}[!ht]
\centering
\includegraphics[width=.82\textwidth]{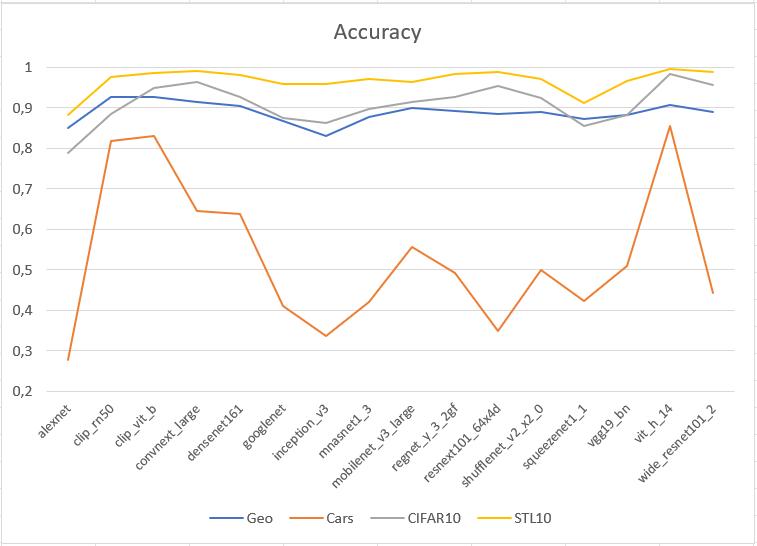}
\caption{Line chart representing the accuracy of each model on each dataset.}
\label{fig:glmacc}
\end{figure}

\begin{figure}[!ht]
\centering
\includegraphics[width=.82\textwidth]{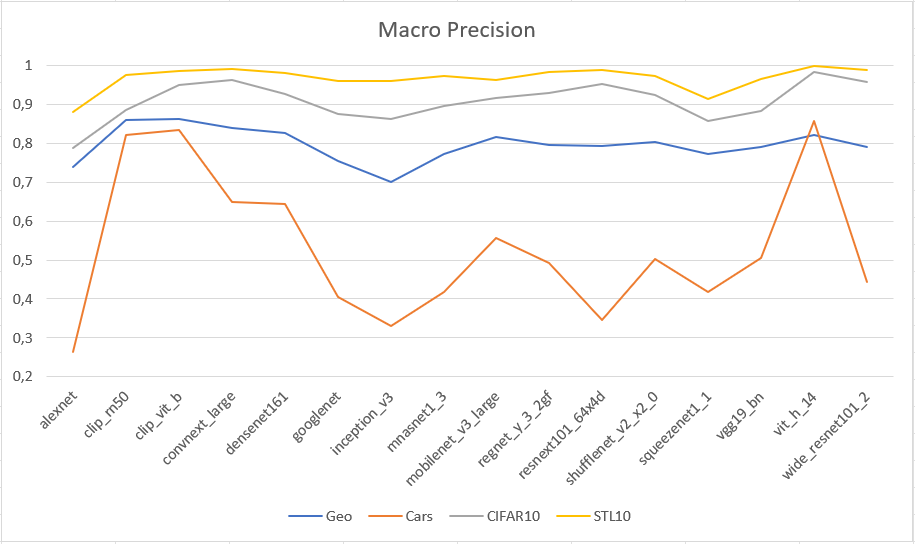}
\caption{Line chart representing the macro precision of each model on each dataset.}
\label{fig:glmmacpre}
\end{figure}

\begin{figure}[!ht]
\centering
\includegraphics[width=.82\textwidth]{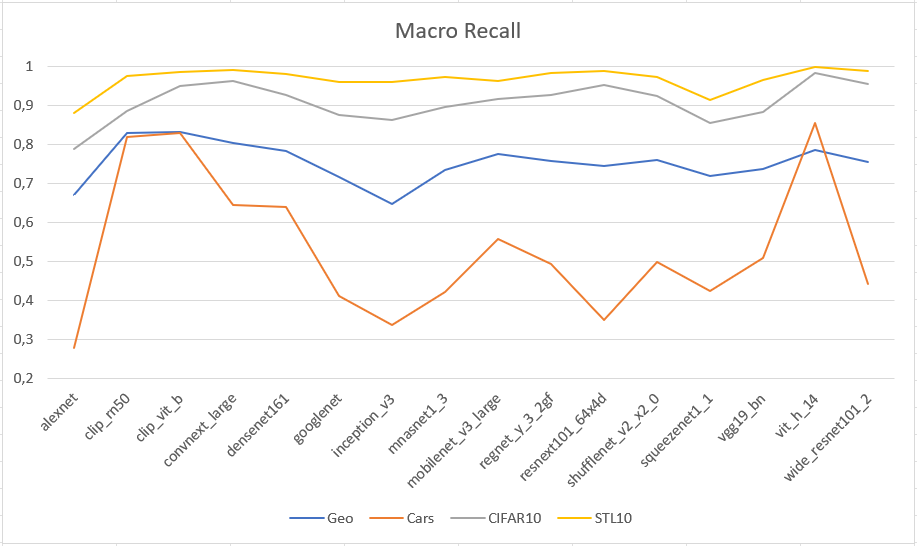}
\caption{Line chart representing the macro recall of each model on each dataset.}
\label{fig:glmmacrec}
\end{figure}

\begin{figure}[!ht]
\centering
\includegraphics[width=.82\textwidth]{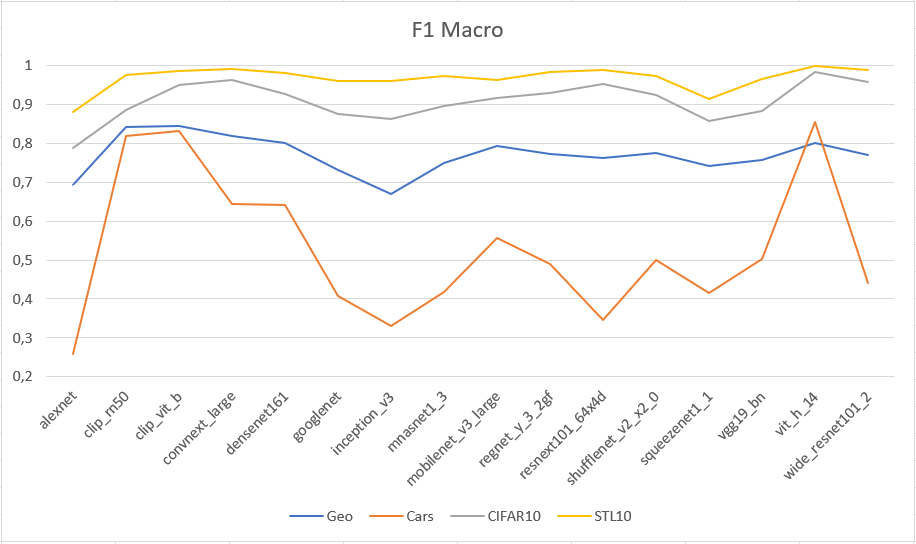}
\caption{Line chart representing the macro f1-score of each model on each dataset.}
\label{fig:glmmacfsco}
\end{figure}

\begin{figure}[!ht]
\centering
\includegraphics[width=.82\textwidth]{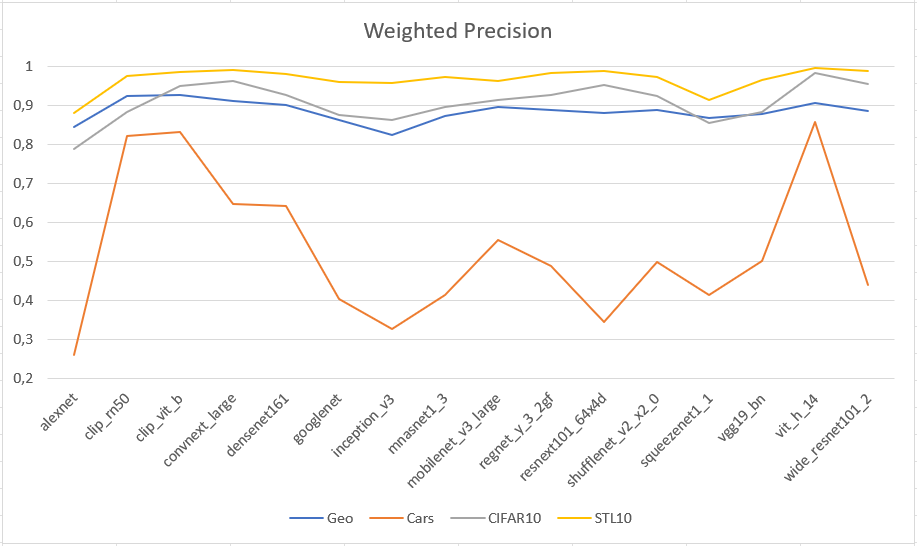}
\caption{Line chart representing the weighted precision of each model on each dataset.}
\label{fig:glmweipre}
\end{figure}

\begin{figure}[!ht]
\centering
\includegraphics[width=.82\textwidth]{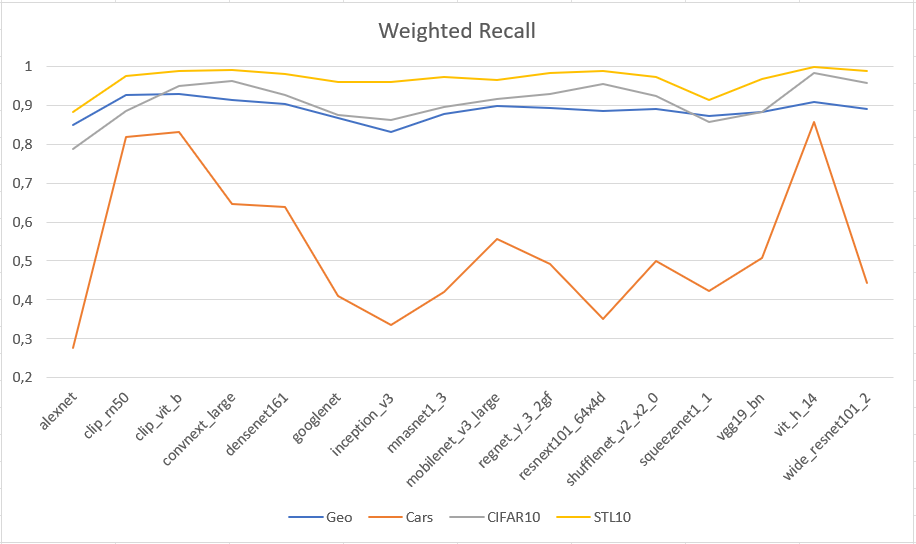}
\caption{Line chart representing the weighted recall of each model on each dataset.}
\label{fig:glmweirec}
\end{figure}

\begin{figure}[!ht]
\centering
\includegraphics[width=.82\textwidth]{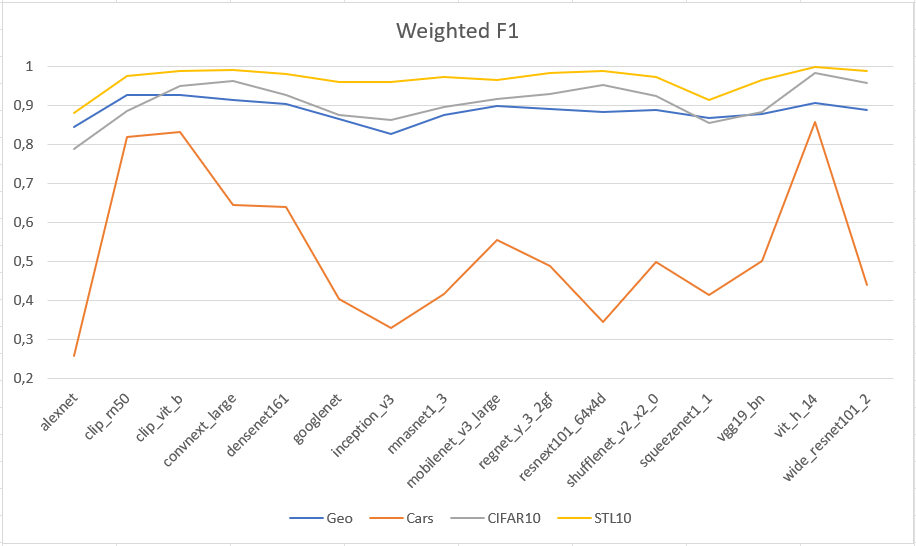}
\caption{Line chart representing the weighted f1-score of each model on each dataset.}
\label{fig:glmweifsco}
\end{figure}

In order to facilitate the data analysis, we have represented the data generated in our experiments in the following line charts, which demonstrate the performance (according to different metrics) of each pre-trained model for classifying the images in the four selected datasets. Figure \ref{fig:glmacc} represents the accuracy. Figure \ref{fig:glmmacpre} shows the macro precision. Figure \ref{fig:glmmacrec} indicates the macro recall. Figure \ref{fig:glmmacfsco} demonstrates the macro f1-score. Figure \ref{fig:glmweipre} presents the weighted precision. Figure \ref{fig:glmweirec} shows the weighted recall. Figure \ref{fig:glmweifsco} indicates the weighted f1-score of each model on each dataset. 

The line charts in Figures \ref{fig:glmmacpre}-\ref{fig:glmweifsco} present a very similar pattern of variation of the model's performance across all datasets. We can notice also that, in general, the model's performance pattern increases and the differences among patterns decrease (resulting in a smoother pattern) in the Geological Images dataset when we consider the accuracy and the weighted averages of precision, recall, and f-score in comparison with the macro averages of these measures. This behavior is expected since the imbalance of this dataset is more pronounced.
We can notice that, in general, the CLIP-ViT-B and VisionTransformer-H/14 models tend to show the best performances, considering all metrics in most datasets. In the Geological Images dataset, the CLIP-ViT-B presents the best performance, in all metrics. The CLIP-ResNet50 also tends to perform well in the other datasets, although in the case of the CIFAR-10 and Geological Images datasets, this model's performance is reasonably lower than the performance of CLIP-ViT-B and ViT-H/14, in all metrics. In the CIFAR-10 dataset, it is also worth highlighting the good performance of ConvNeXt Large and ResNeXt101-64x4D. The ConvNeXt Large model also performs better than CLIP-ViT-B and CLIP-ResNet50 in STL10.
In our evaluation, AlexNet presents the worst performance on most datasets, considering all metrics. Inception V3 also performed poorly on the Stanford Cars, CIFAR-10, and Geological Images datasets, where it performed the worst. Another model that had reasonably low performance compared to the others was Squezenet1-1. The poor performance of this model is more pronounced on the CIFAR-10 and STL10 datasets.

\begin{figure}[!ht]
\centering
\includegraphics[width=.82\textwidth]{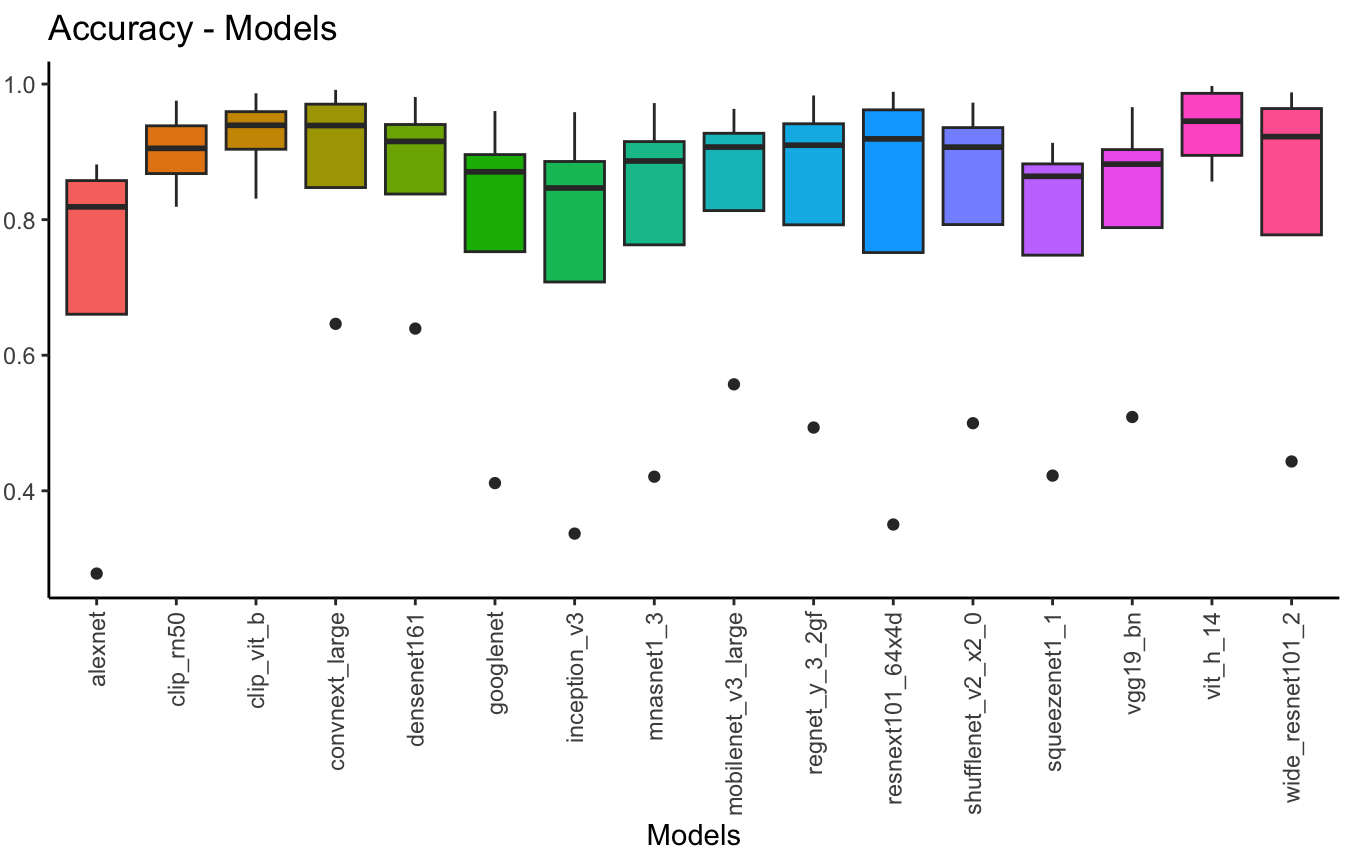}
\caption{Boxplot of accuracy for each model.}
\label{fig:boxmodacc}
\end{figure}

\begin{figure}[!ht]
\centering
\includegraphics[width=.82\textwidth]{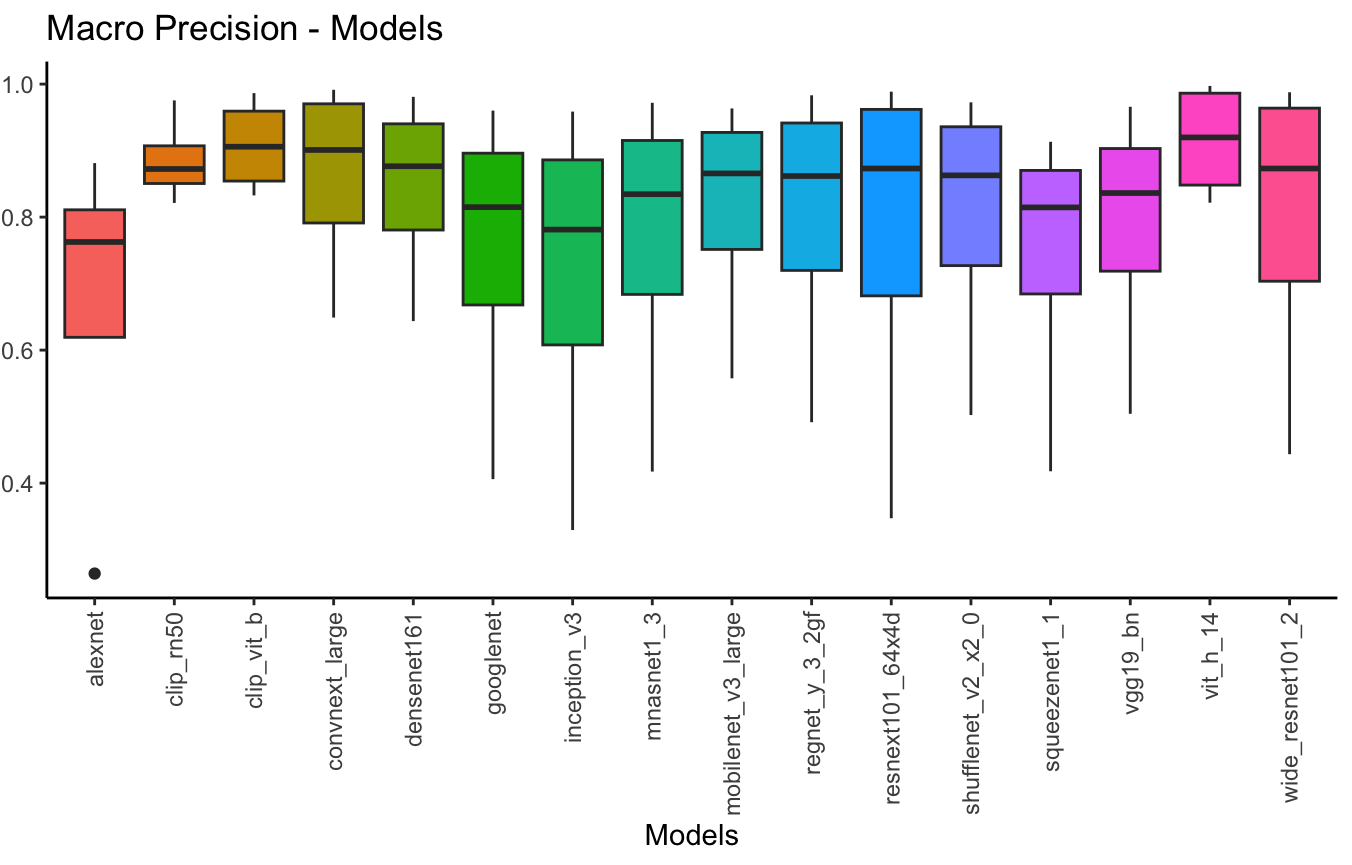}
\caption{Boxplot of macro precision for each model.}
\label{fig:boxmodmacpre}
\end{figure}

\begin{figure}[!ht]
\centering
\includegraphics[width=.82\textwidth]{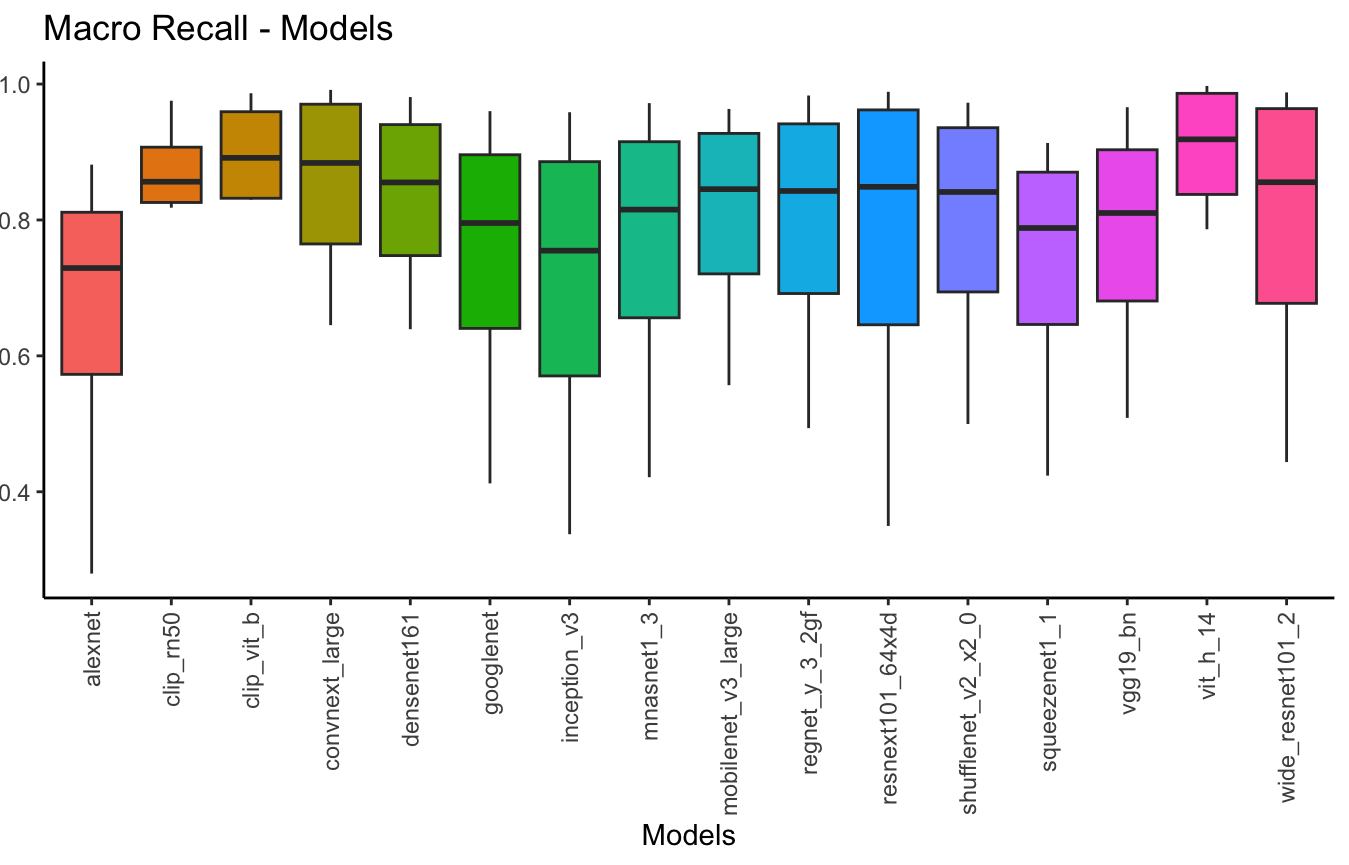}
\caption{Boxplot of macro recall for each model.}
\label{fig:boxmodmacrec}
\end{figure}

\begin{figure}[!ht]
\centering
\includegraphics[width=.82\textwidth]{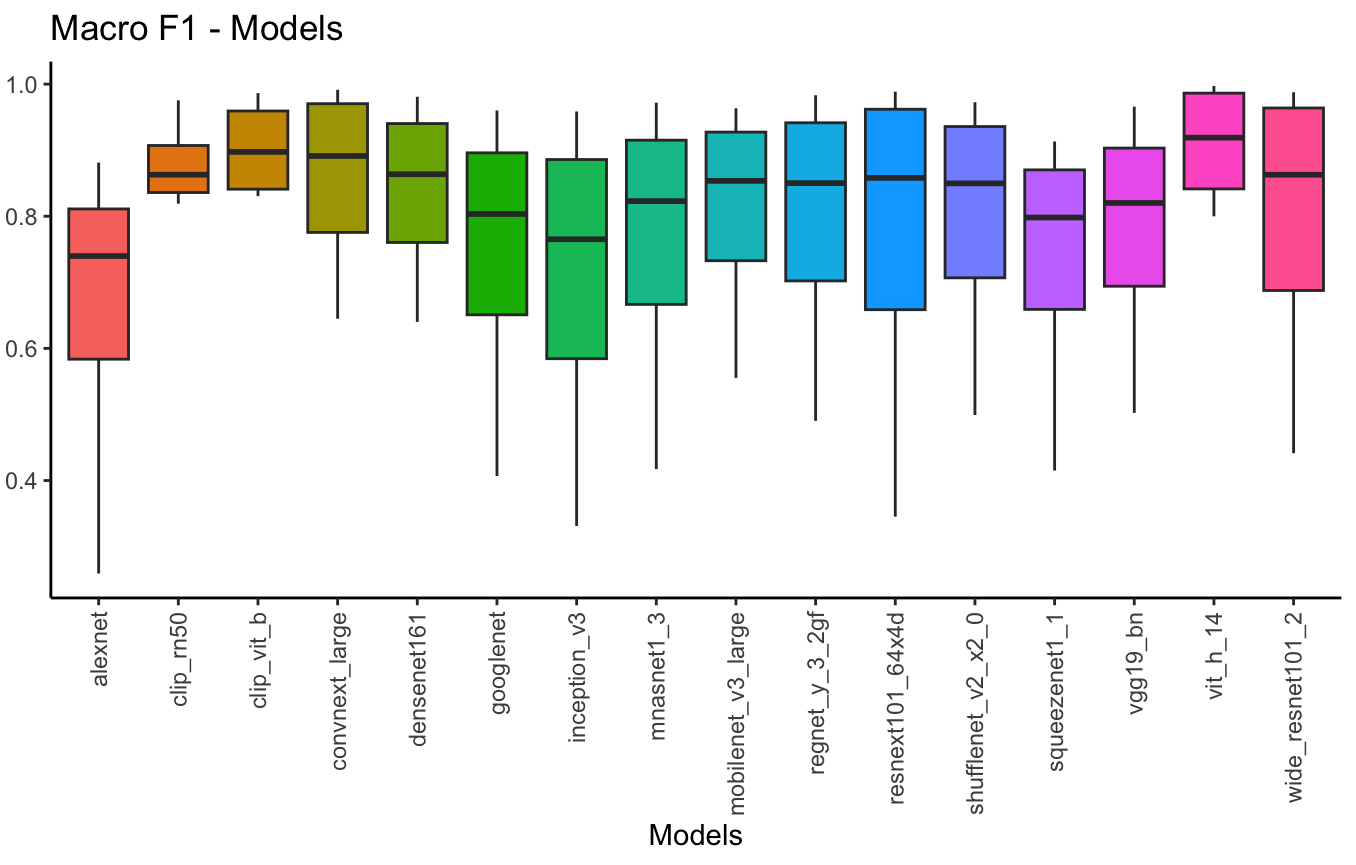}
\caption{Boxplot of f1-score for each model.}
\label{fig:boxmodmacfsco}
\end{figure}

\begin{figure}[!ht]
\centering
\includegraphics[width=.82\textwidth]{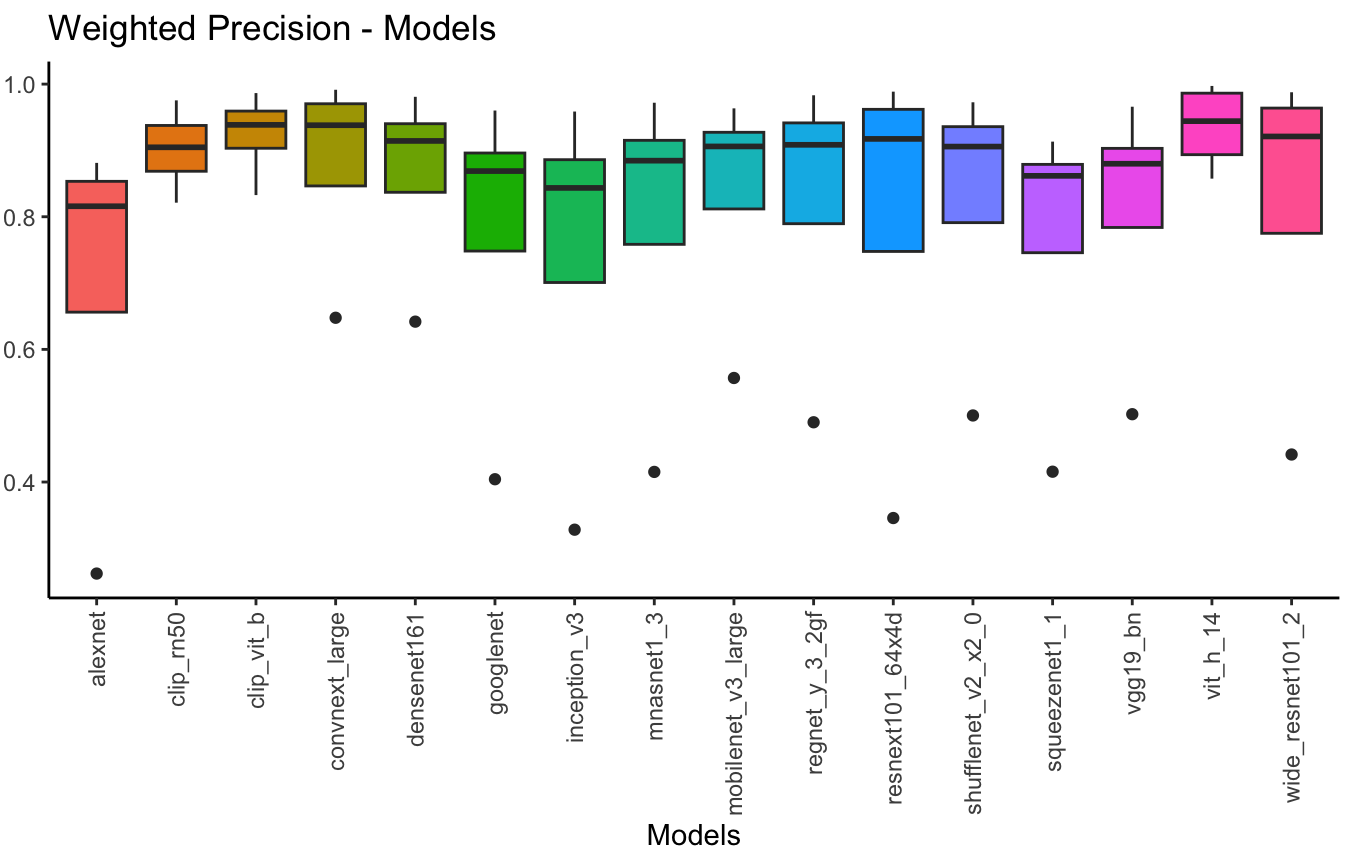}
\caption{Boxplot of weighted precision for each model.}
\label{fig:boxmodweipre}
\end{figure}

\begin{figure}[!ht]
\centering
\includegraphics[width=.82\textwidth]{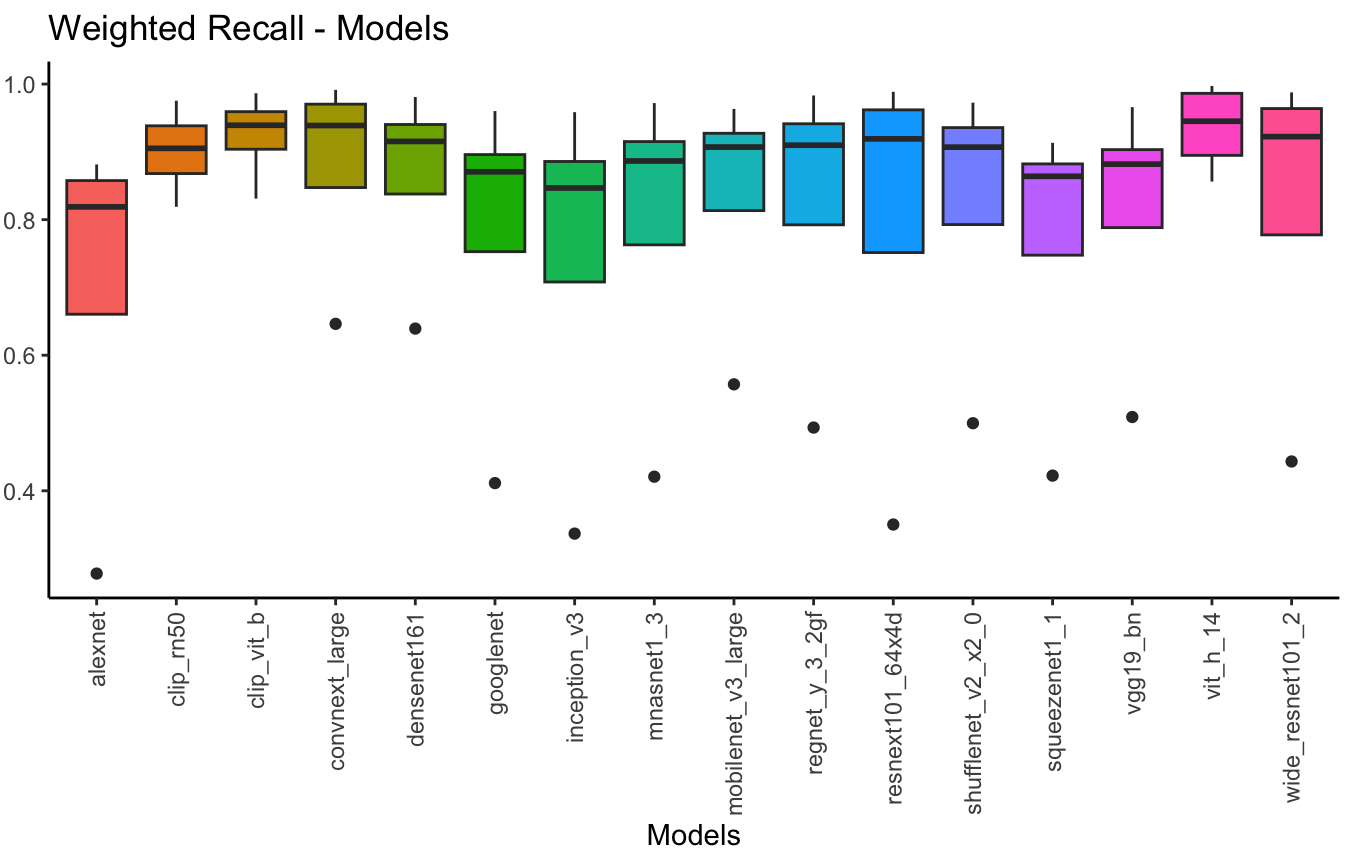}
\caption{Boxplot of weighted recall for each model.}
\label{fig:boxmodweirec}
\end{figure}

\begin{figure}[!ht]
\centering
\includegraphics[width=.82\textwidth]{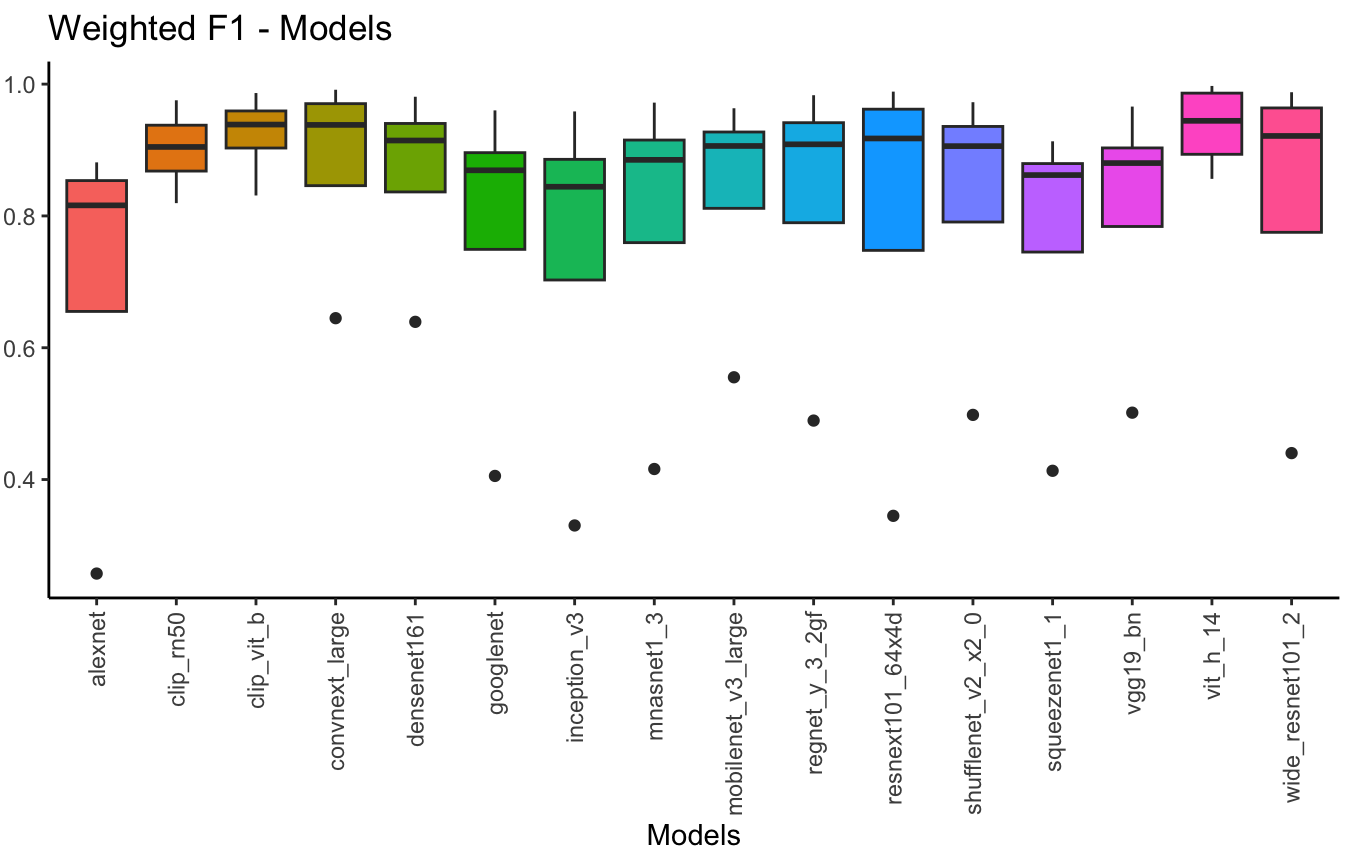}
\caption{Boxplot of weighted f1-score for each model.}
\label{fig:boxmodweifsco}
\end{figure}

The following boxplots reveal the minimum value (excluding outliers), the maximum value (excluding outliers), the median, the first quartile, the third quartile, and outliers of the performance of the different models considering all the datasets, according to different metrics. Figure \ref{fig:boxmodacc} represents the accuracy. Figure \ref{fig:boxmodmacpre} shows the macro precision. Figure \ref{fig:boxmodmacrec} indicates the macro recall. Figure \ref{fig:boxmodmacfsco} demonstrates the macro f1-score. Figure \ref{fig:boxmodweipre} presents the weighted precision. Figure \ref{fig:boxmodweirec} represents the weighted recall. Figure \ref{fig:boxmodweifsco} shows the weighted f1-score of each model on each dataset. Notice that each boxplot is built from only four measures (one for each dataset), however, despite this small amount of data, this visualization can provide further insights.

In the boxplots represented in Figures \ref{fig:boxmodacc}-\ref{fig:boxmodweifsco}, it is possible to notice a pattern in the different metrics, although some differences can be noticed in the Geological Images dataset when comparing the macro averages of precision, recall, and f-score with the accuracy and weighted averages of those metrics.
The CLIP-ResNet50, CLIP-ViT-B, and VisionTransformer-H/14 models present the lowest variability in performance, in general. In the macro averages of precision, recall, and f-score, CLIP-ResNet50 presents the lowest variability. On the other hand, when considering the accuracy and weighted averages of precision, recall, and f-score, CLIP-ViT-B presents less variability in its performance. It is important to note that CLIP-ResNet50, CLIP-ViT-B, and VisionTransformer-H/14 are the only models adopted in our experiments that include transformers in their architecture.
It is also worth noting that, among the CNN-based architectures, ConvNeXt Large presented a higher median, compared to the other CNN-based architectures, which is similar to the medians of performances obtained by CLIP-ViT-B and VisionTransformer-H/14, although with greater variability in performance. The boxplots also suggest that, in the considered datasets, AlexNet tends to present the worst performances, in general, presenting low medians and high variability. High variability is also present in the performances of ResNeXt101-64x4D, Wide ResNet 101-2, and Inception V3, which also presents low medians when compared to the other models.

\begin{figure}[!ht]
\centering
\includegraphics[width=1\textwidth]{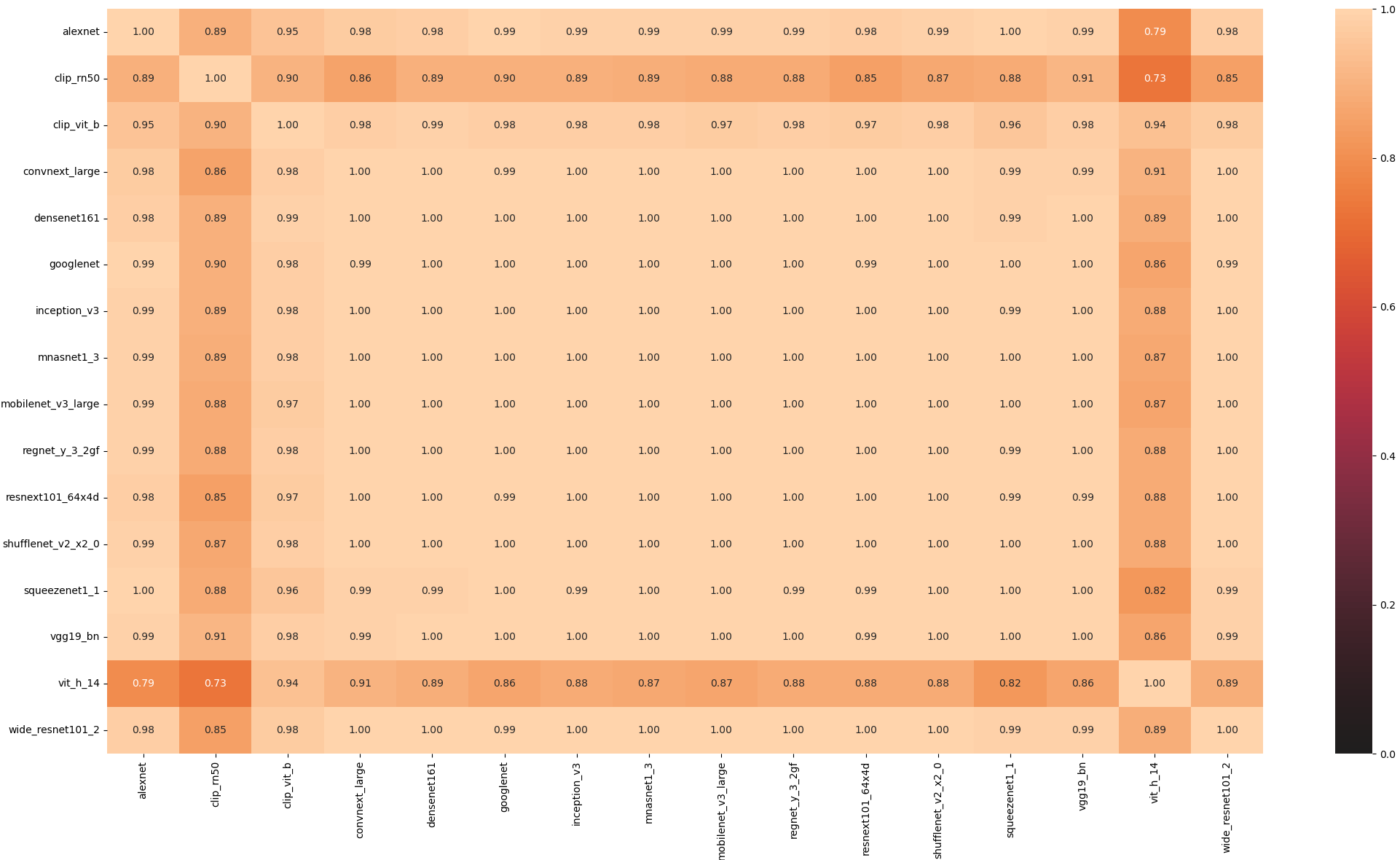}
\caption{Heat map representing the correlation between each pair of models regarding accuracy.}
\label{fig:heatacc}
\end{figure}

\begin{figure}[!ht]
\centering
\includegraphics[width=1\textwidth]{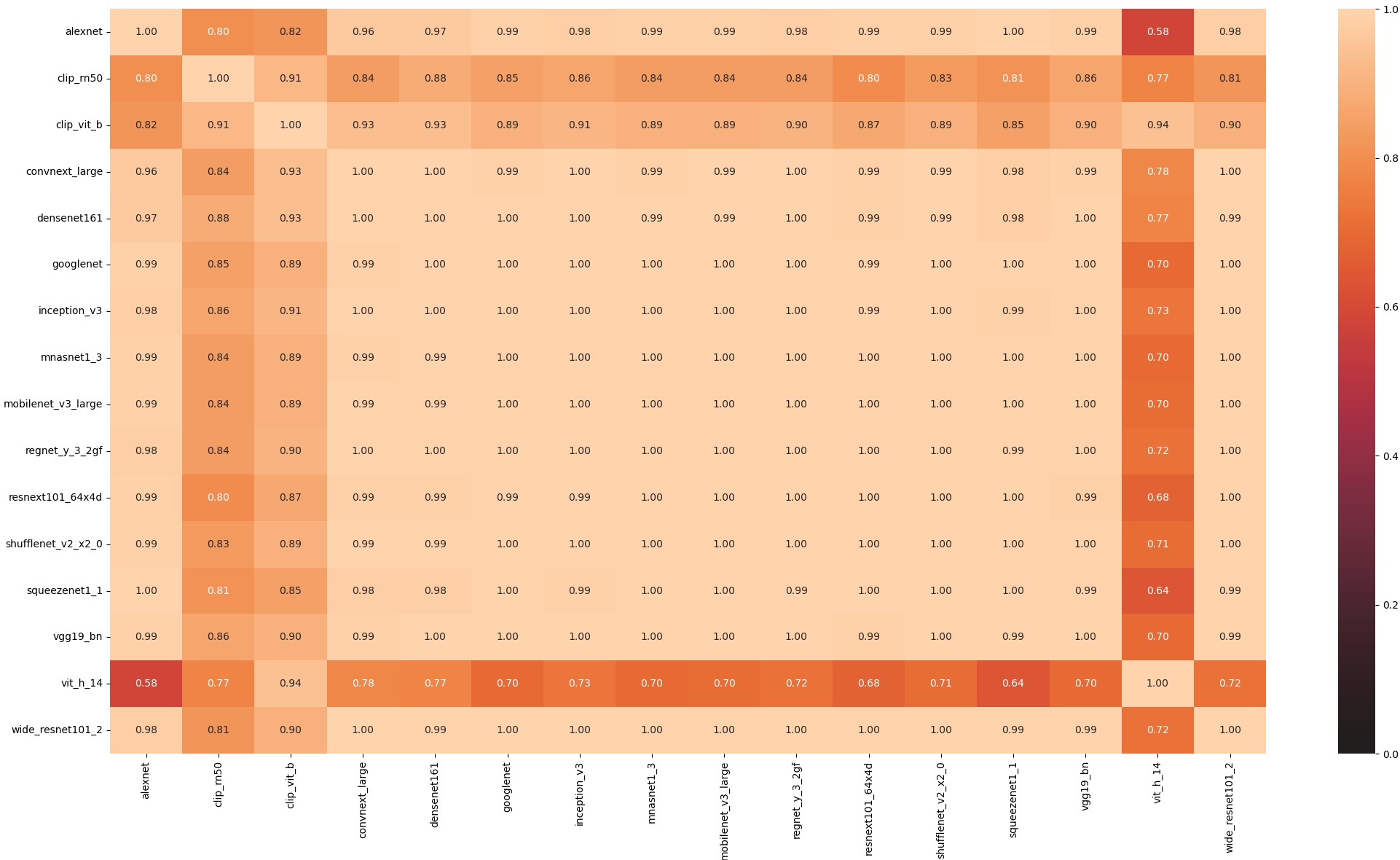}
\caption{Heat map representing the correlation between each pair of models regarding the macro precision.}
\label{fig:heatmacpre}
\end{figure}

\begin{figure}[!ht]
\centering
\includegraphics[width=1\textwidth]{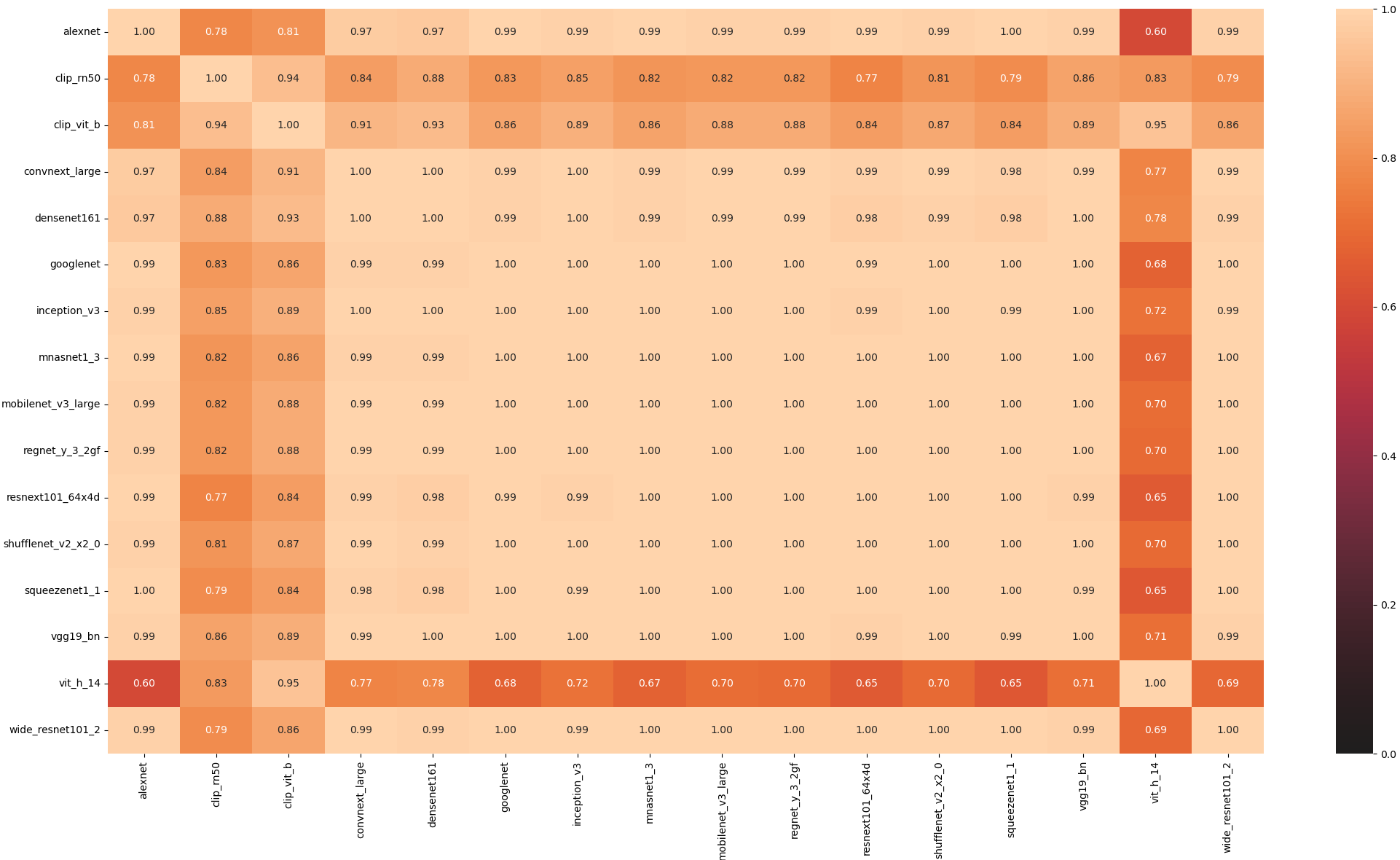}
\caption{Heat map representing the correlation between each pair of models regarding the macro recall.}
\label{fig:heatmacrec}
\end{figure}

\begin{figure}[!ht]
\centering
\includegraphics[width=1\textwidth]{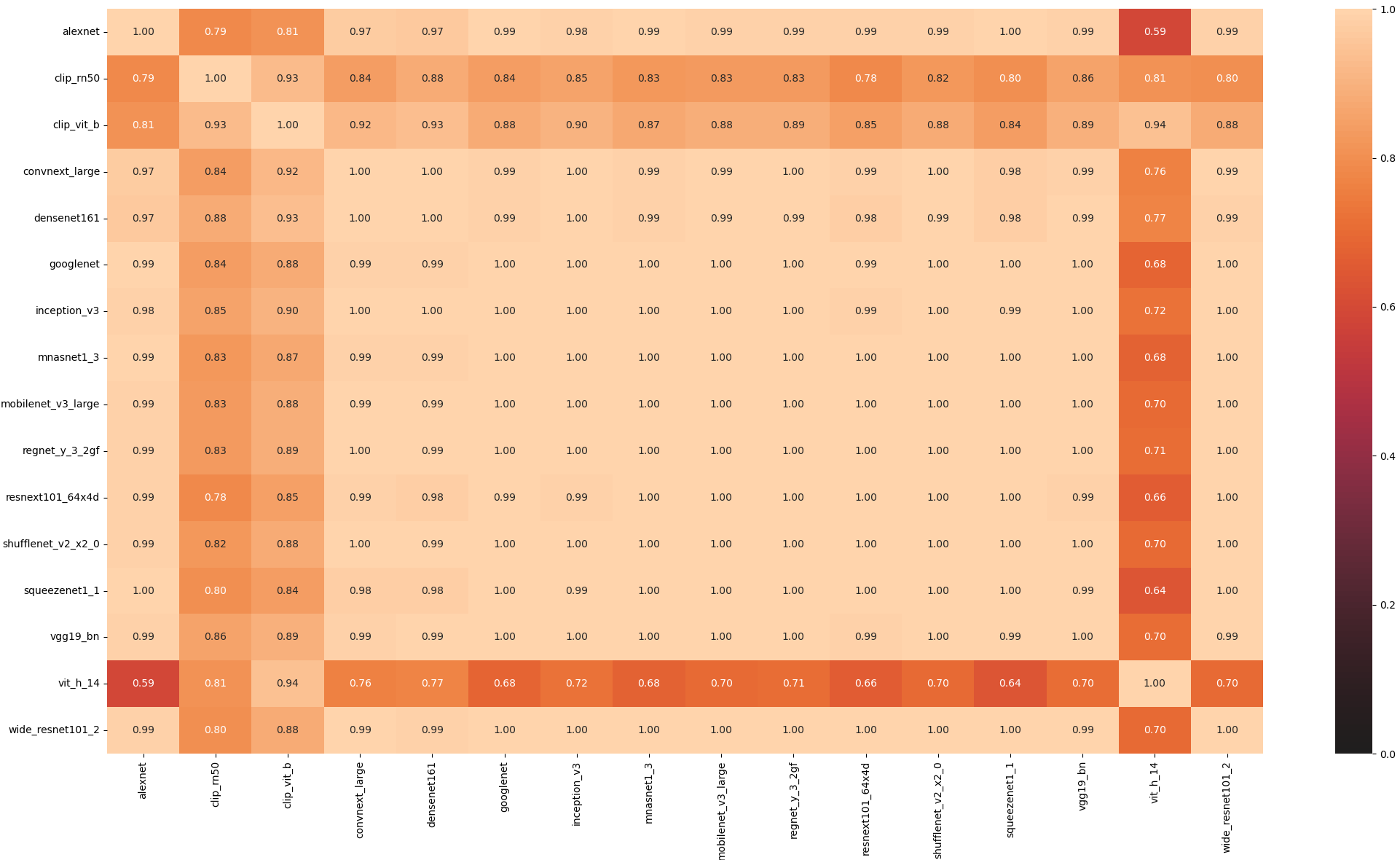}
\caption{Heat map representing the correlation between each pair of models regarding the macro f1-score.}
\label{fig:heatmacfsco}
\end{figure}

\begin{figure}[!ht]
\centering
\includegraphics[width=1\textwidth]{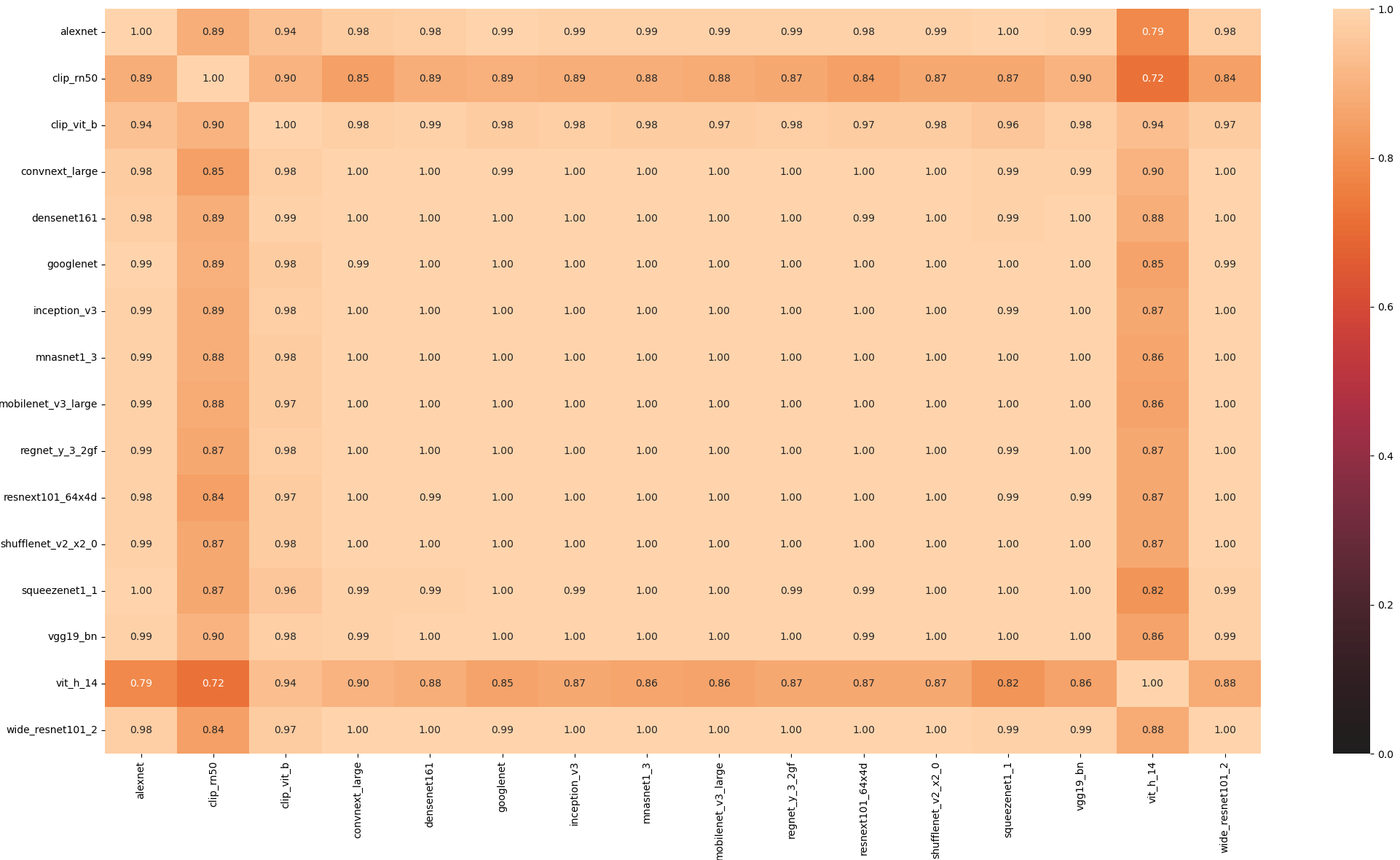}
\caption{Heat map representing the correlation between each pair of models regarding the weighted precision.}
\label{fig:heatweipre}
\end{figure}

\begin{figure}[!ht]
\centering
\includegraphics[width=1\textwidth]{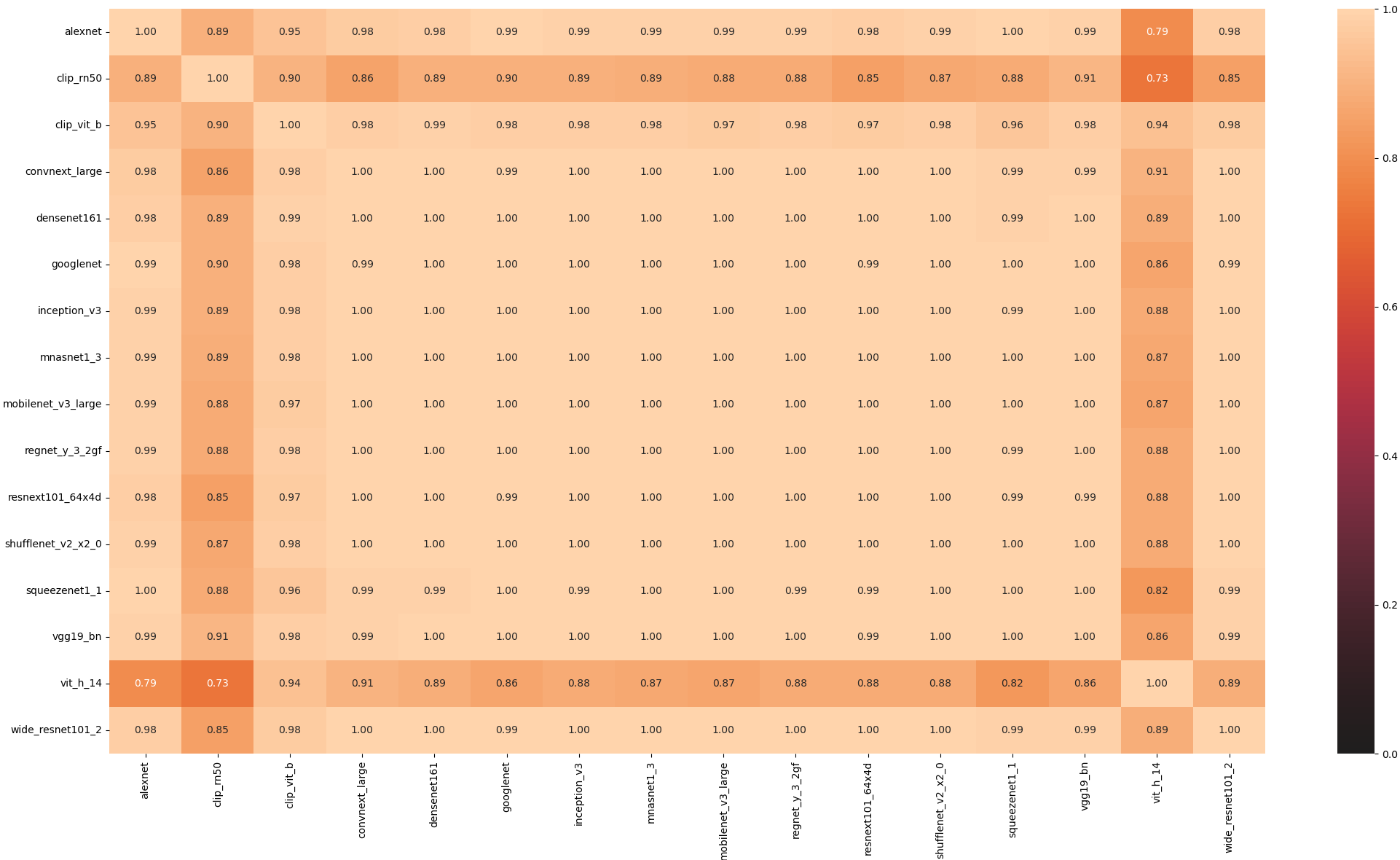}
\caption{Heat map representing the correlation between each pair of models regarding the weighted recall.}
\label{fig:heatweirec}
\end{figure}

\begin{figure}[!ht]
\centering
\includegraphics[width=1\textwidth]{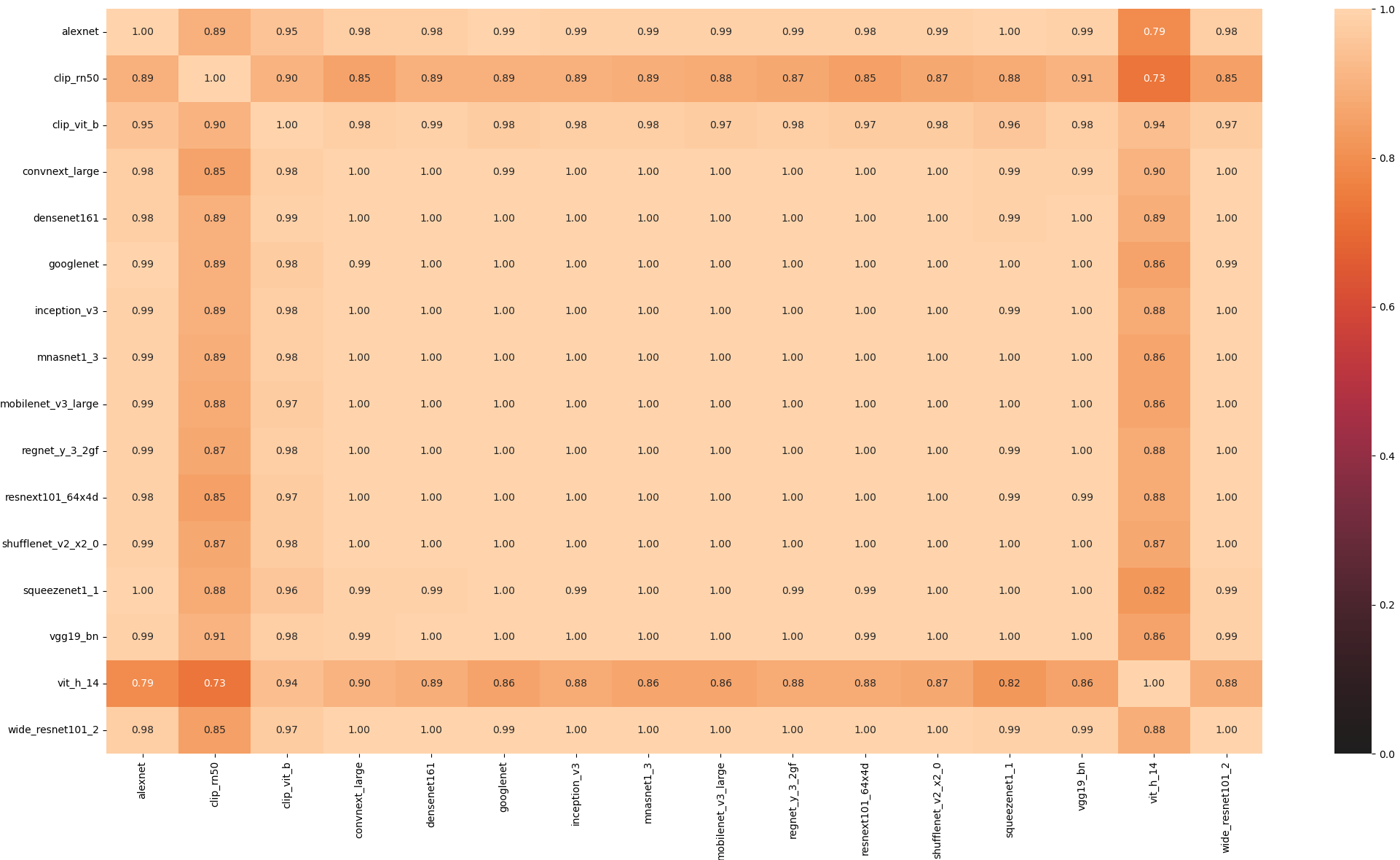}
\caption{Heat map representing the correlation between each pair of models regarding the weighted f1-score.}
\label{fig:heatweifsco}
\end{figure}

The previous analysis (Figures \ref{fig:glmacc}-\ref{fig:glmweifsco}) suggests that some models present a very similar performance behavior across the datasets while other models exhibit behaviors that do not follow the general pattern. In order to emphasize how similar are the model's behaviors, we analyzed the Pearson correlation \cite{cohen:09} of the performances of each pair of models across the datasets according to all the selected metrics. The following heat maps were created to visually represent this information. In the following sequence of charts, the darker a cell gets corresponds to the lower the correlation of a given pair of models, according to a given performance metric. Figure \ref{fig:heatacc} represents the correlation regarding the accuracy. Figure \ref{fig:heatmacpre} shows the correlation regarding the macro precision. Figure \ref{fig:heatmacrec} indicates the correlation regarding the macro recall. Figure \ref{fig:heatmacfsco} demonstrates the correlation regarding the macro f1-score. Figure \ref{fig:heatweipre} presents the correlation regarding the weighted precision. Figure \ref{fig:heatweirec} represents the correlation regarding the weighted recall. Figure \ref{fig:heatweifsco} shows the correlation regarding the weighted f1-score between each pair of models.

Figures \ref{fig:heatacc}-\ref{fig:heatweifsco} suggest that the correlation of the performances of each pair of models presents a similar pattern in all metrics. We can notice also that, in all performance metrics, the correlation between models based solely on CNN architectures is high (in general, above 0.97). However, the performances of CLIP-ResNet50, CLIP-ViT-B, and VisionTransformer-H/14 models present a lower correlation with the performances of other models solely based on CNN. In the case of CLIP-ViT-B, the correlation with the other models is subtly lower considering accuracy and the weighted averages of precision, recall, and f-score. However, when we consider the macro averages of precision, recall, and f-score, this model's correlation is significantly lower. It is important to note that the CLIP-ResNet50, CLIP-ViT-B, and VisionTransformer-H/14 models include transformers in their architectures. This performance correlation analysis suggests that this difference in the basic principles of the architecture of these models is correlated with this difference in the performance pattern of these models when compared to architectures based solely on CNN. Further analysis should be done in the future in order to investigate this hypothesis. The heat maps also allow us to note that the correlations among the performances of CLIP-ResNet50, CLIP-ViT-B, and VisionTransformer-H/14 models are not high when compared with the correlations among the performances of models based solely on CNN.

\begin{figure}[!ht]
\centering
\includegraphics[width=.82\textwidth]{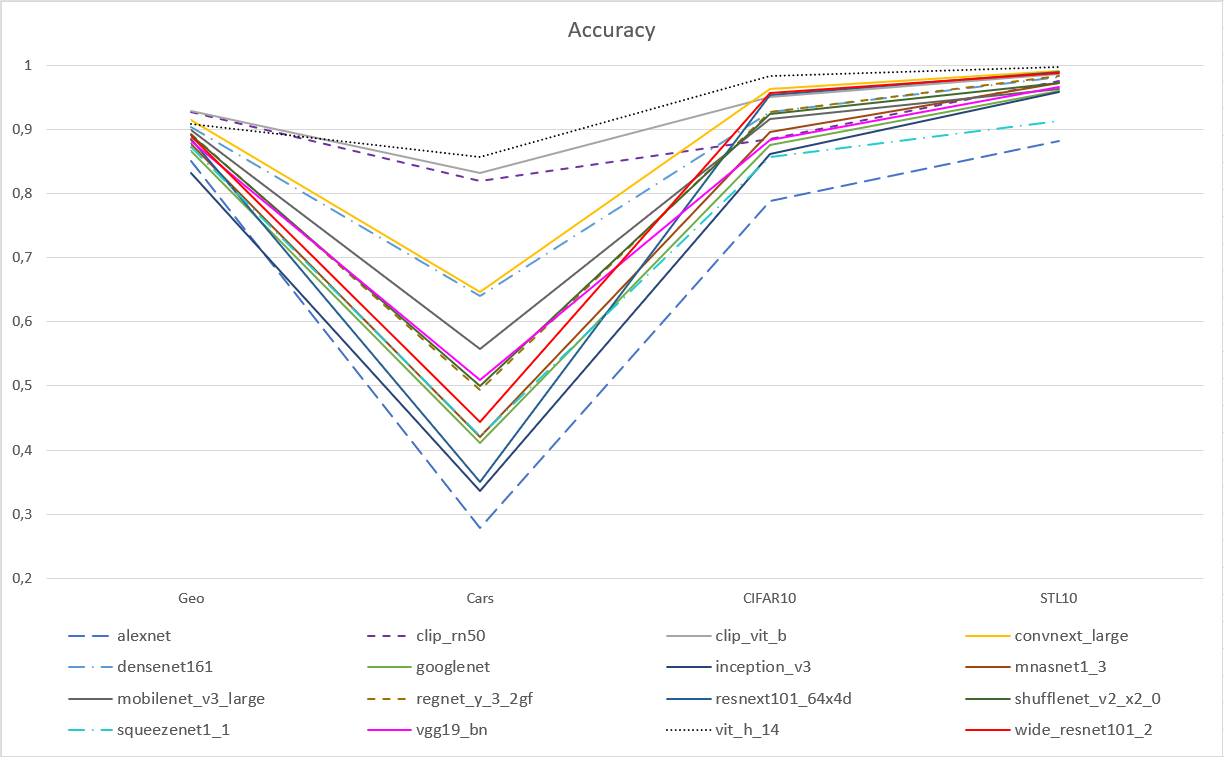}
\caption{Chart representing the accuracy of all models on different datasets.}
\label{fig:gldatacc}
\end{figure}

\begin{figure}[!ht]
\centering
\includegraphics[width=.82\textwidth]{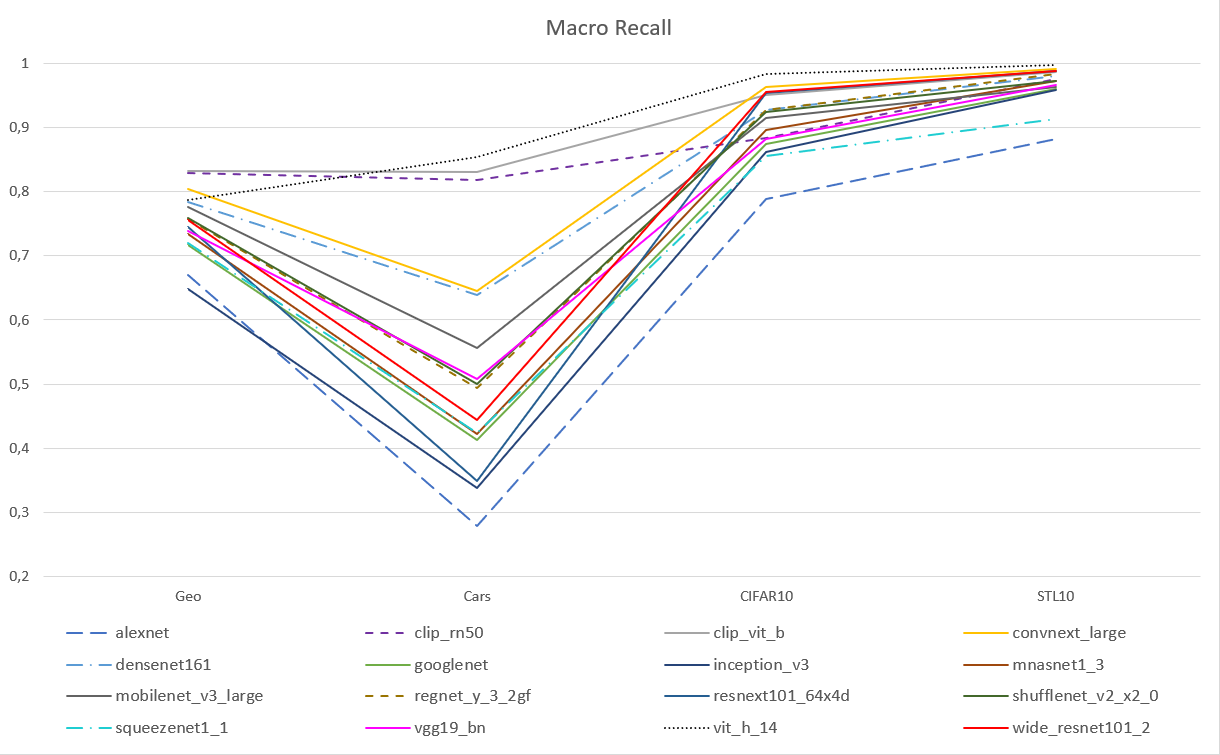}
\caption{Chart representing the macro recall of all models on different datasets.}
\label{fig:gldatmacrec}
\end{figure}

\begin{figure}[!ht]
\centering
\includegraphics[width=.82\textwidth]{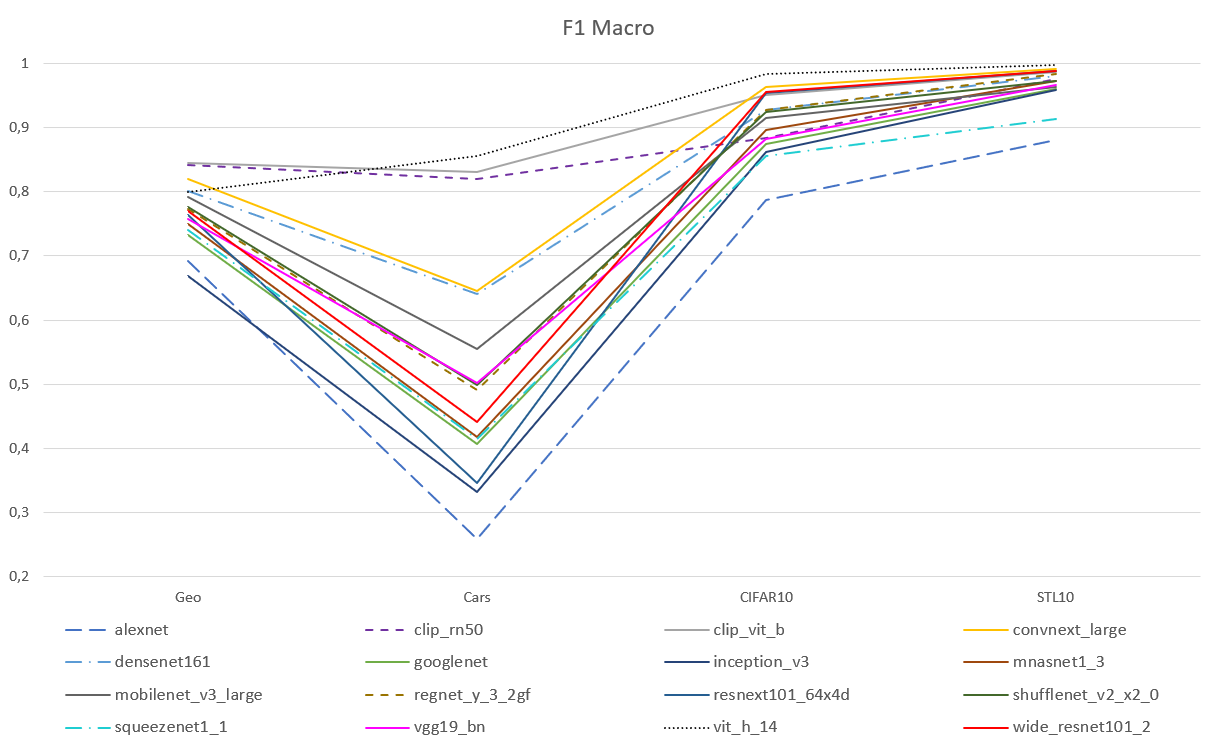}
\caption{Chart representing the macro f1-score of all models on different datasets.}
\label{fig:gldatmacfsco}
\end{figure}

\begin{figure}[!ht]
\centering
\includegraphics[width=.82\textwidth]{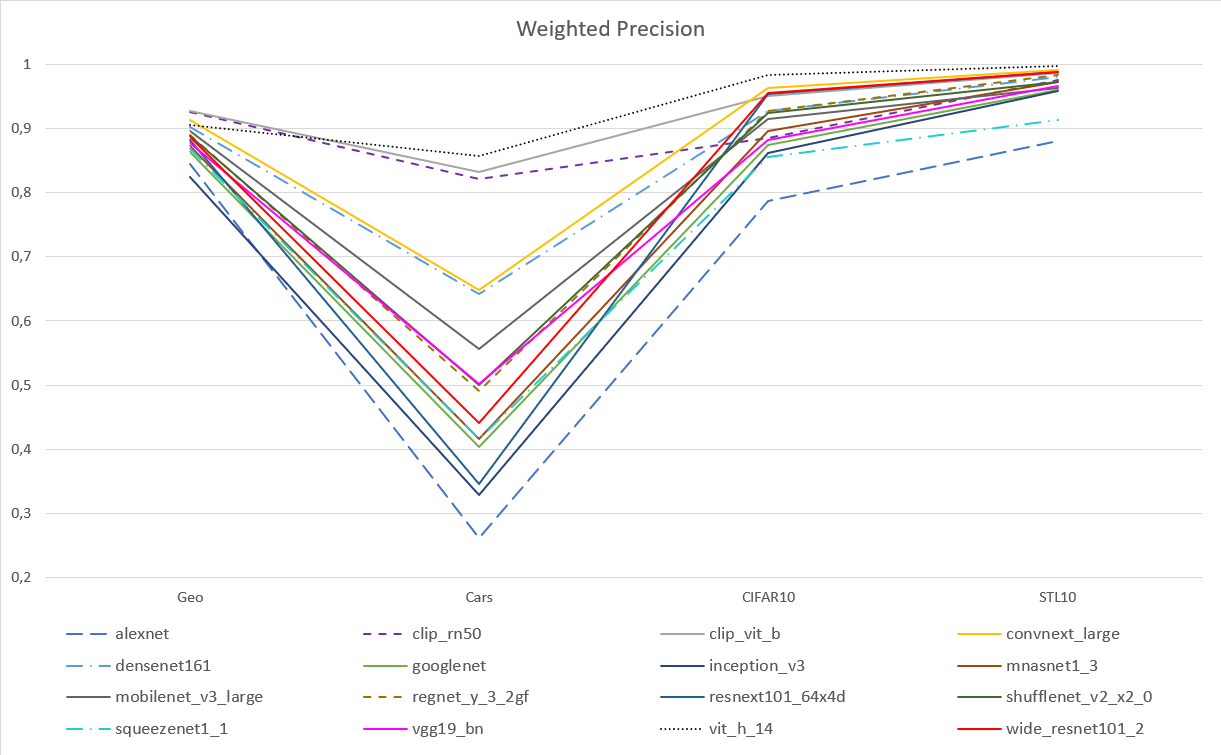}
\caption{Chart representing the weighted precision of all models on different datasets.}
\label{fig:gldatweipre}
\end{figure}

\begin{figure}[!ht]
\centering
\includegraphics[width=.82\textwidth]{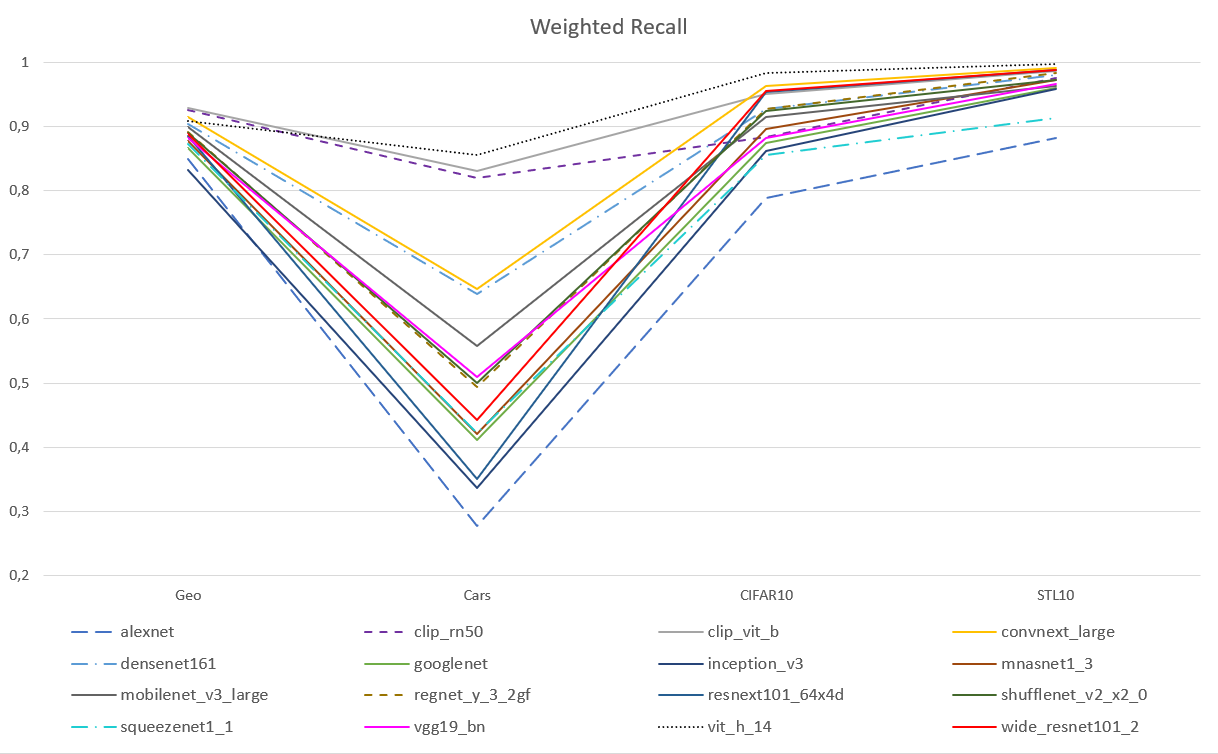}
\caption{Chart representing the weighted recall of all models on different datasets.}
\label{fig:gldatweirec}
\end{figure}

\begin{figure}[!ht]
\centering
\includegraphics[width=.82\textwidth]{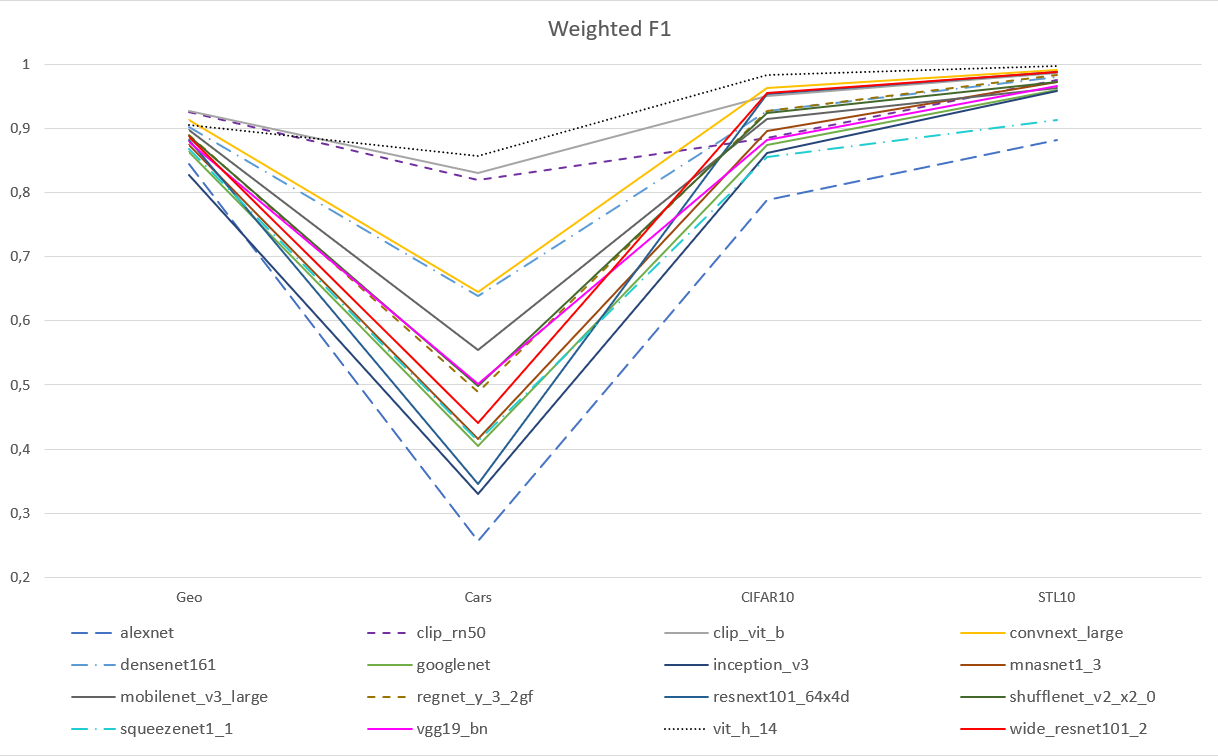}
\caption{Chart representing the weighted f1-score of all models on different datasets.}
\label{fig:gldatweifsco}
\end{figure}

In the previous analyses, we focused on the performance of the models considered in our experiments. In the following charts, we focused on analyzing the datasets considered in our experiments. Each chart represents how the performances of all models vary across the different datasets, considering different metrics. This information is already present in Figures \ref{fig:glmacc} - \ref{fig:glmweifsco}. However, these charts provide an alternative visualization of this information that can suggest, for example, how hard is to classify each dataset. Figure \ref{fig:gldatacc} focuses on accuracy. Figure \ref{fig:glmmacpre} represents macro precision. Figure \ref{fig:gldatmacrec} shows macro recall. Figure \ref{fig:gldatmacfsco} indicates macro f1-score. Figure \ref{fig:gldatweipre} demonstrates weighted precision. Figure \ref{fig:gldatweirec} presents weighted recall. Figure \ref{fig:gldatweifsco} focuses on the weighted f1-score of all models across the datasets.

\begin{figure}[!ht]
\centering
\includegraphics[width=.82\textwidth]{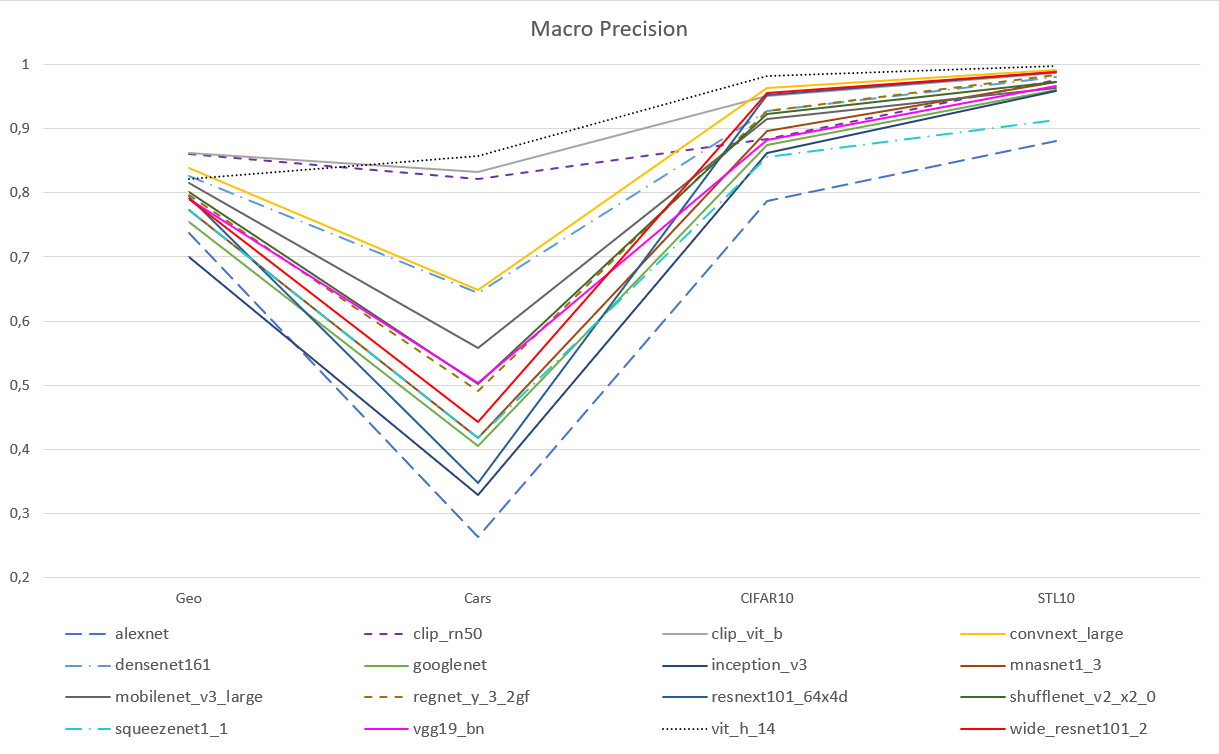}
\caption{Chart representing the macro precision of all models on different datasets.}
\label{fig:gldatmacpre}
\end{figure}

The line charts represented in Figures \ref{fig:gldatacc}-\ref{fig:gldatweifsco} present a similar pattern that can be seen across the different metrics. There are subtle differences when comparing the macro averages of precision, recall, and f-score with accuracy and weighted averages of those metrics. Note that, in general, the models tend to perform better in the STL10 dataset, in second place CIFAR-10 has the best overall results, in third place the dataset of Geological Images and, finally, the dataset with the worst performances, in general, is the Stanford Cars. This is an expected result since this dataset has a large number of classes, few samples per class, and the inclusion of images with different sizes and features at different scales. The Geological Images dataset has similar properties but has fewer classes and more samples per class than Stanford Cars, although it presents a greater imbalance.

\begin{figure}[!ht]
\centering
\includegraphics[width=.82\textwidth]{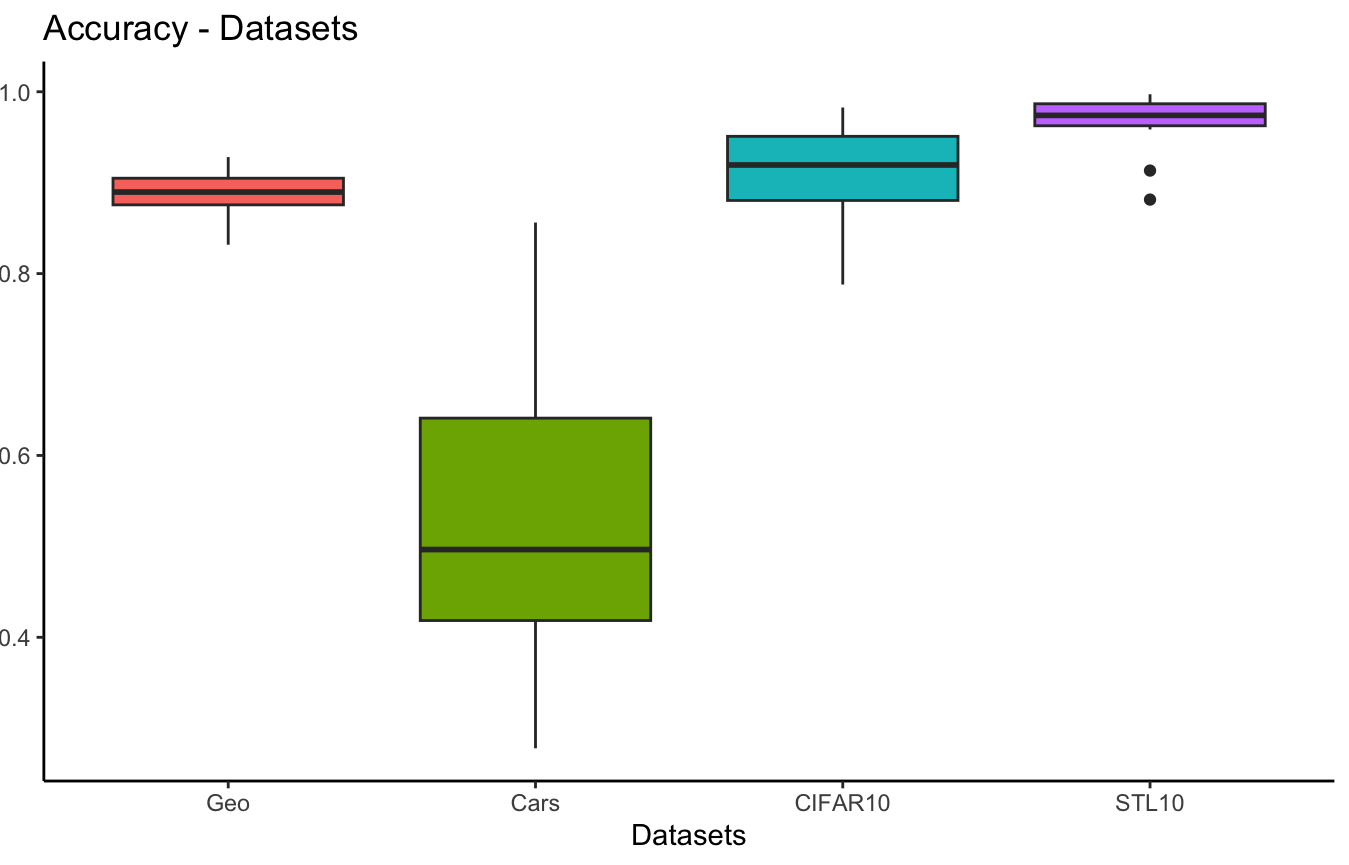}
\caption{Boxplot of accuracy for each dataset.}
\label{fig:boxdatacc}
\end{figure}

\begin{figure}[!ht]
\centering
\includegraphics[width=.82\textwidth]{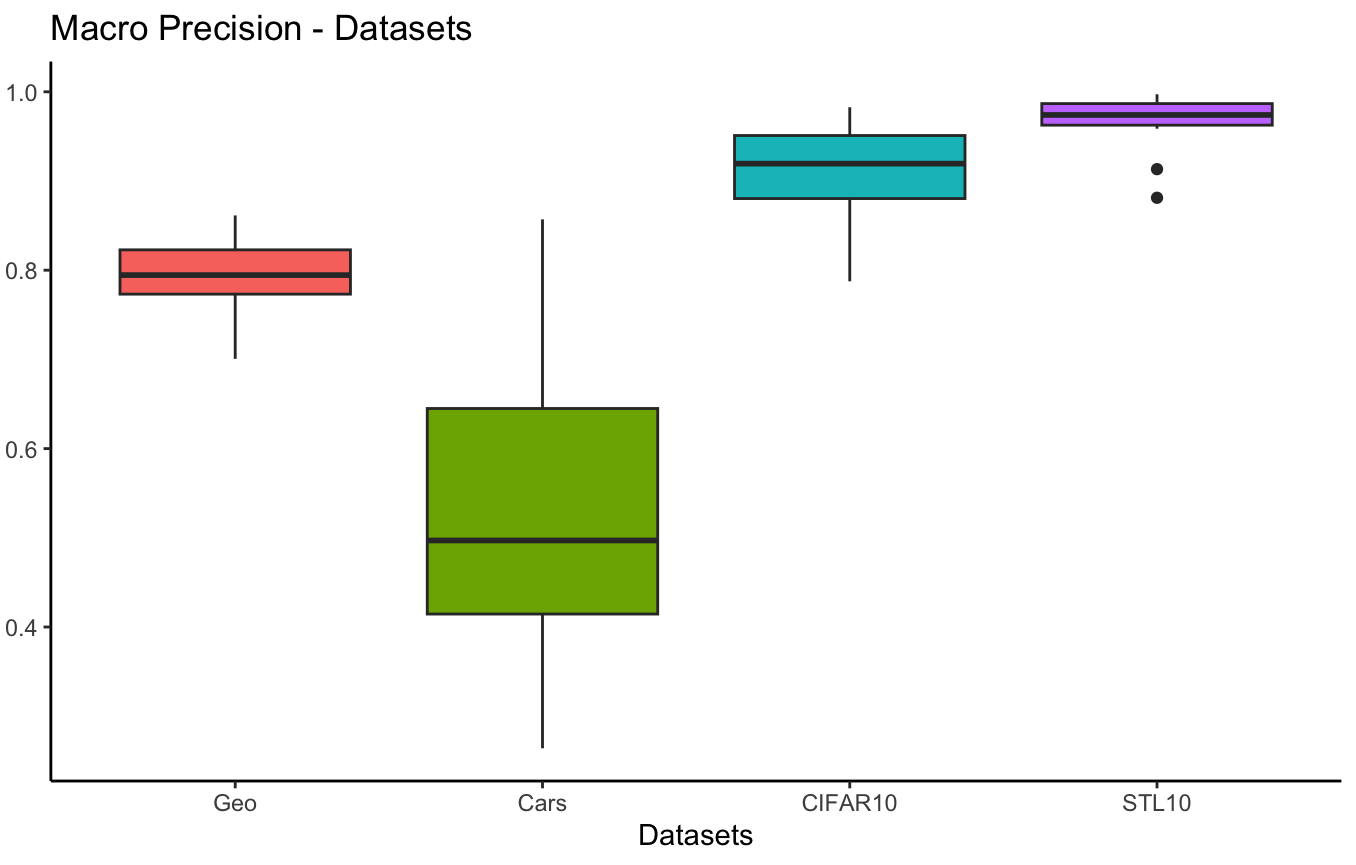}
\caption{Boxplot of macro precision for each dataset.}
\label{fig:boxdatmacpre}
\end{figure}

\begin{figure}[!ht]
\centering
\includegraphics[width=.82\textwidth]{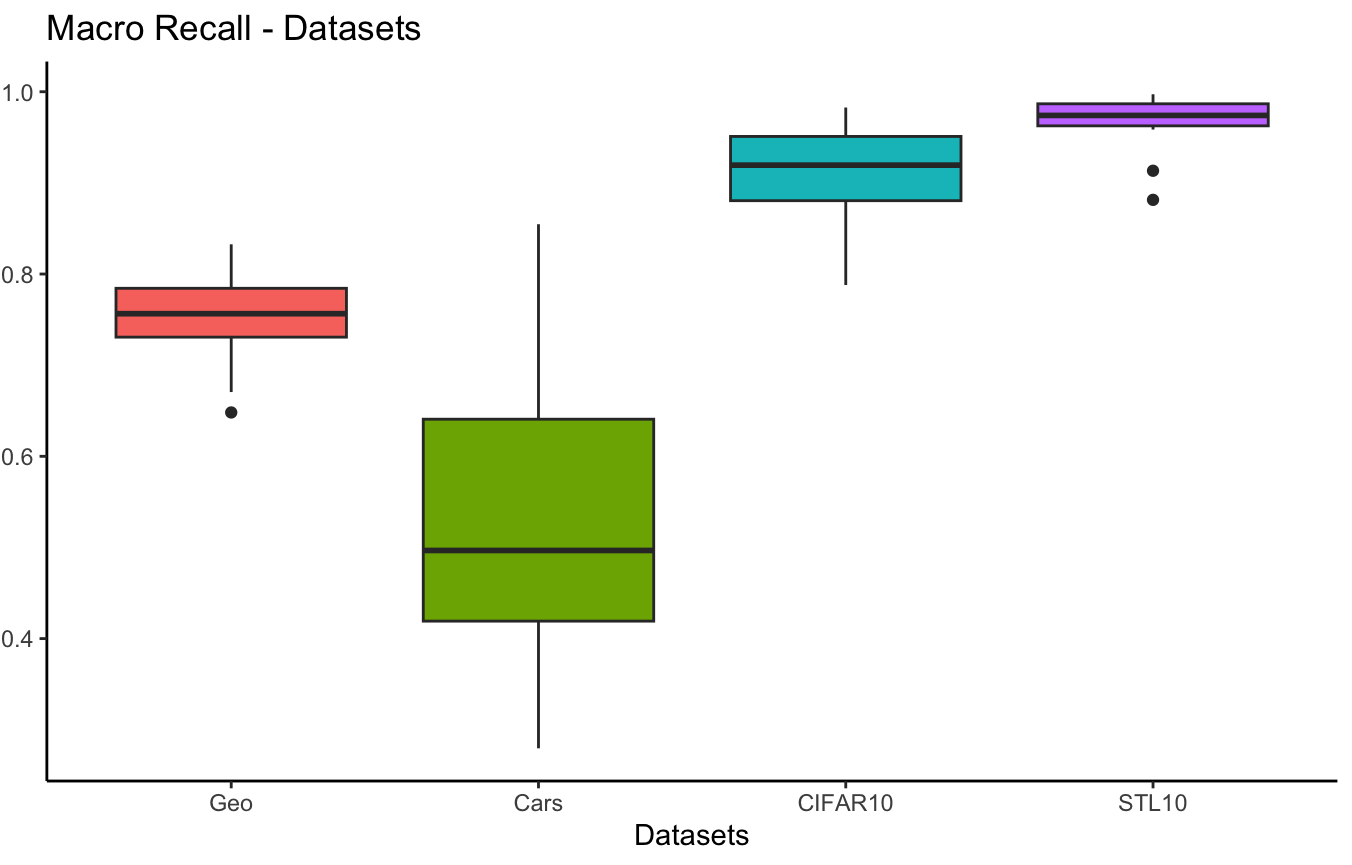}
\caption{Boxplot of macro recall for each dataset.}
\label{fig:boxdatmacrec}
\end{figure}

\begin{figure}[!ht]
\centering
\includegraphics[width=.82\textwidth]{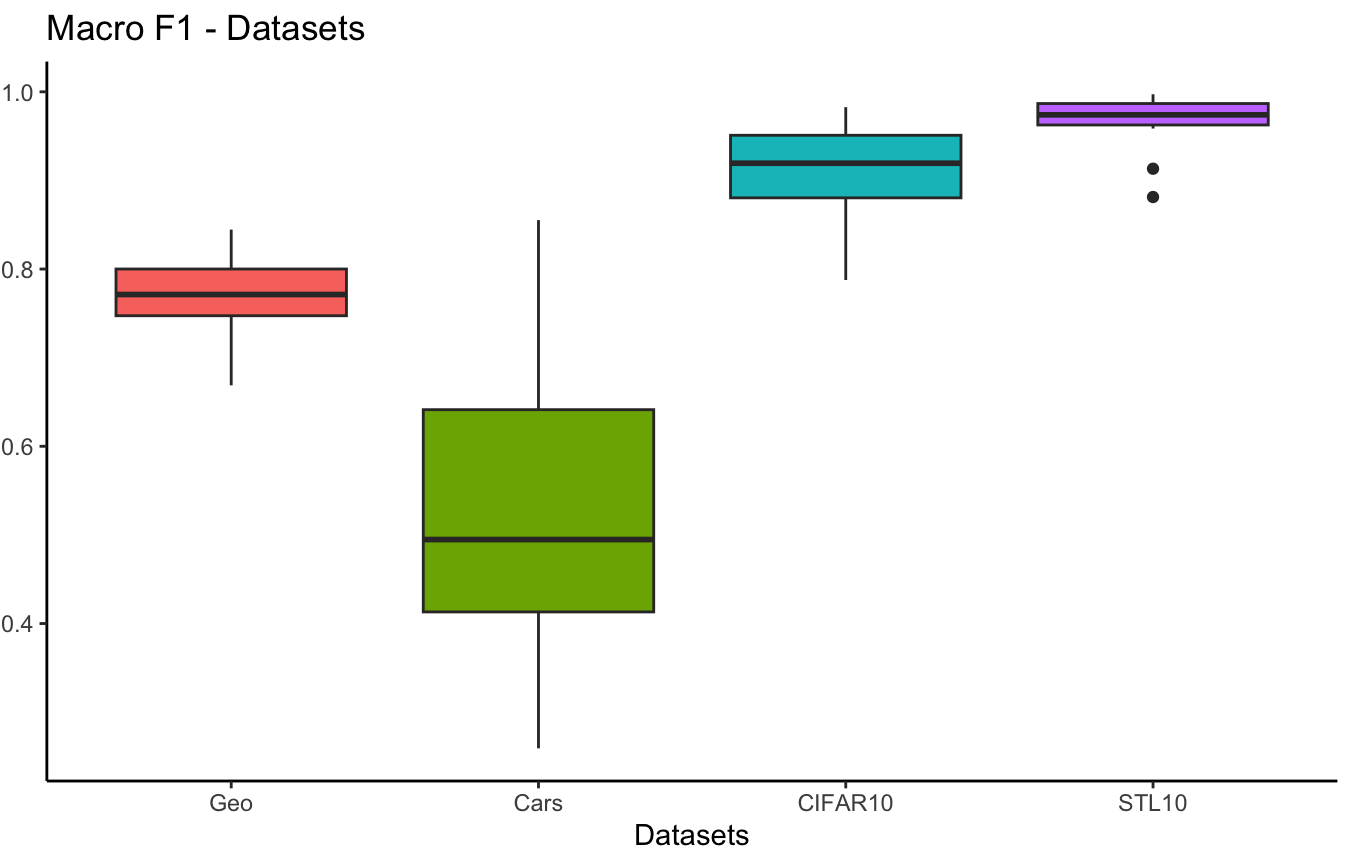}
\caption{Boxplot of macro f1-score for each dataset.}
\label{fig:boxdatmacfsco}
\end{figure}

\begin{figure}[!ht]
\centering
\includegraphics[width=.82\textwidth]{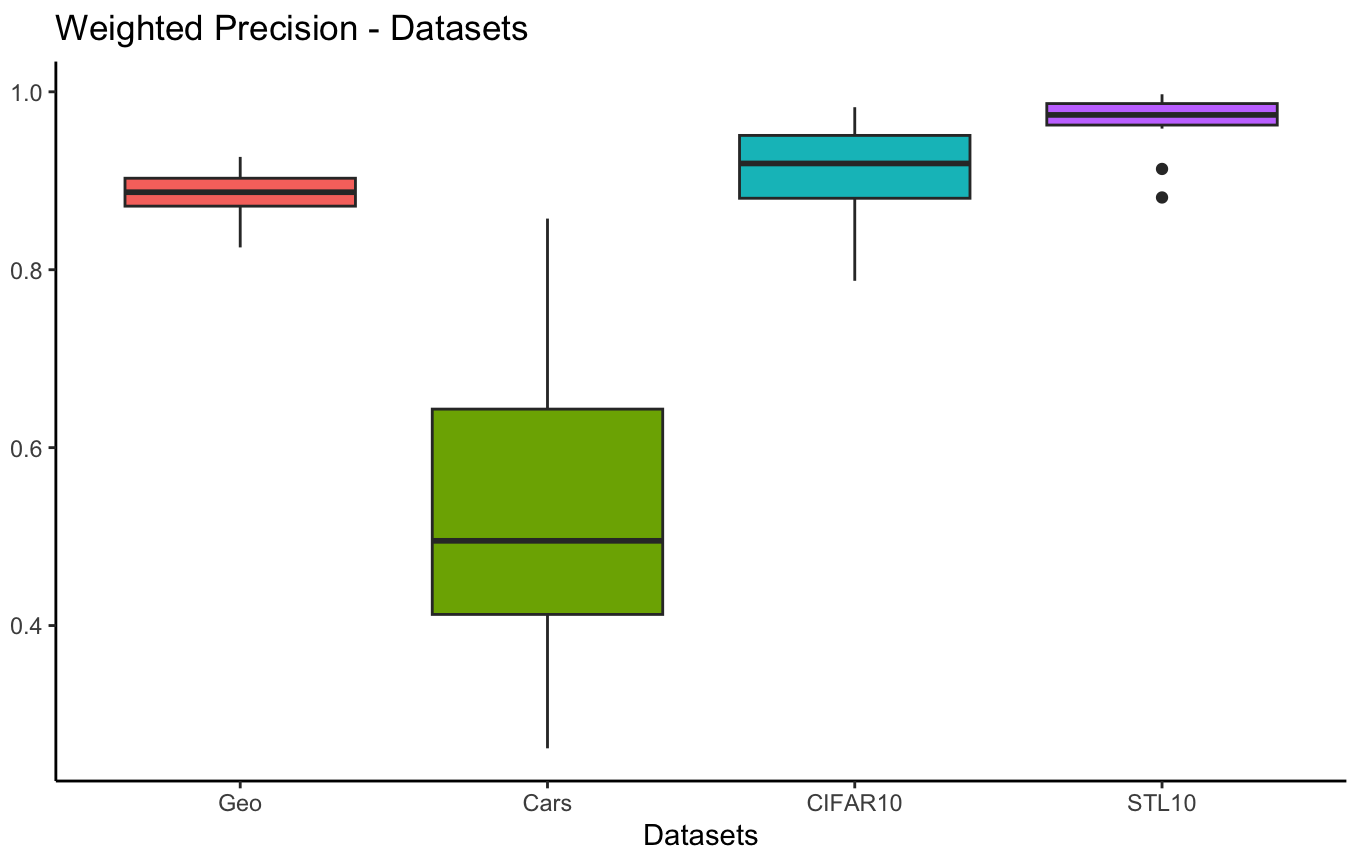}
\caption{Boxplot of weighted precision for each dataset.}
\label{fig:boxdatweipre}
\end{figure}

\begin{figure}[!ht]
\centering
\includegraphics[width=.82\textwidth]{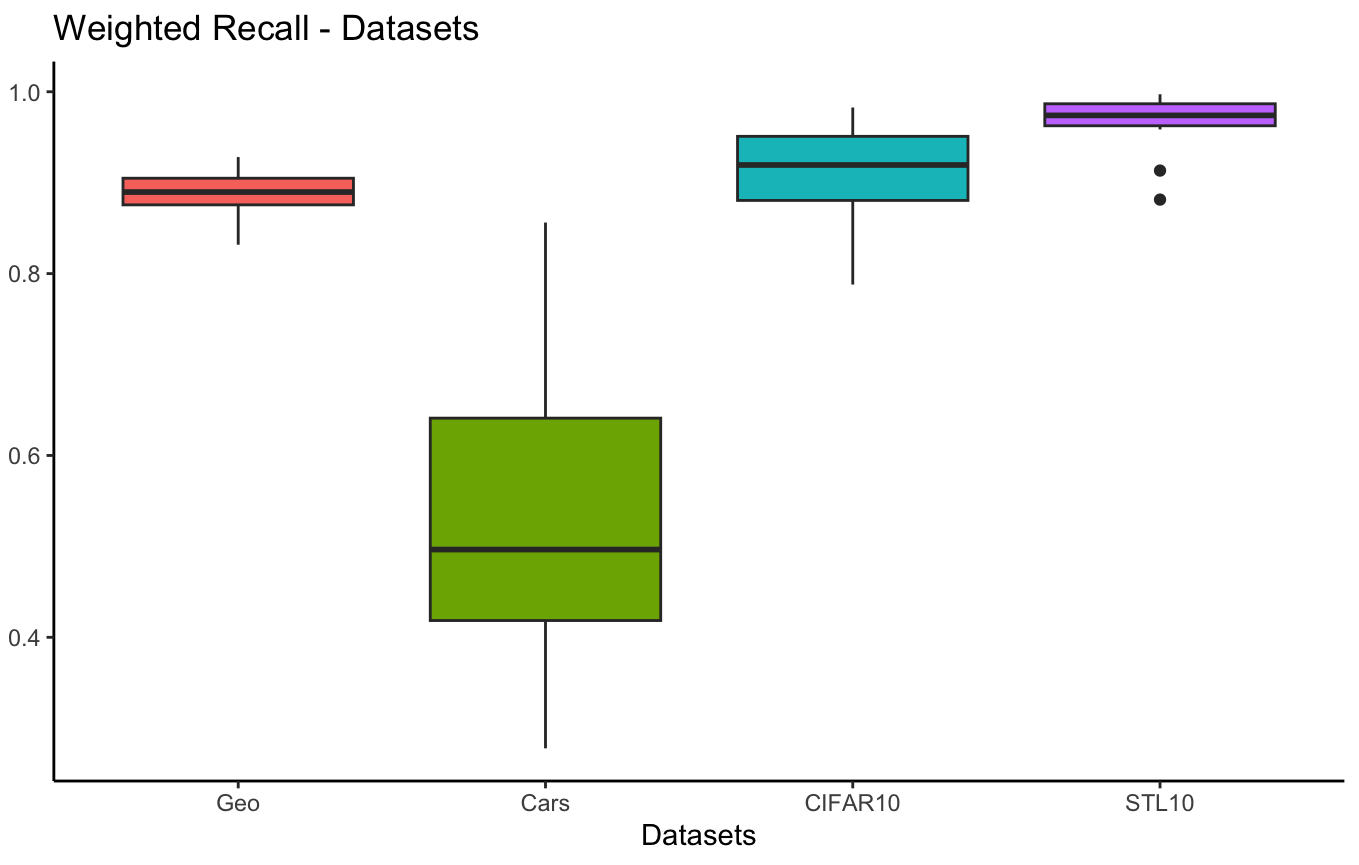}
\caption{Boxplot of weighted recall for each dataset.}
\label{fig:boxdatweirec}
\end{figure}

\begin{figure}[!ht]
\centering
\includegraphics[width=.82\textwidth]{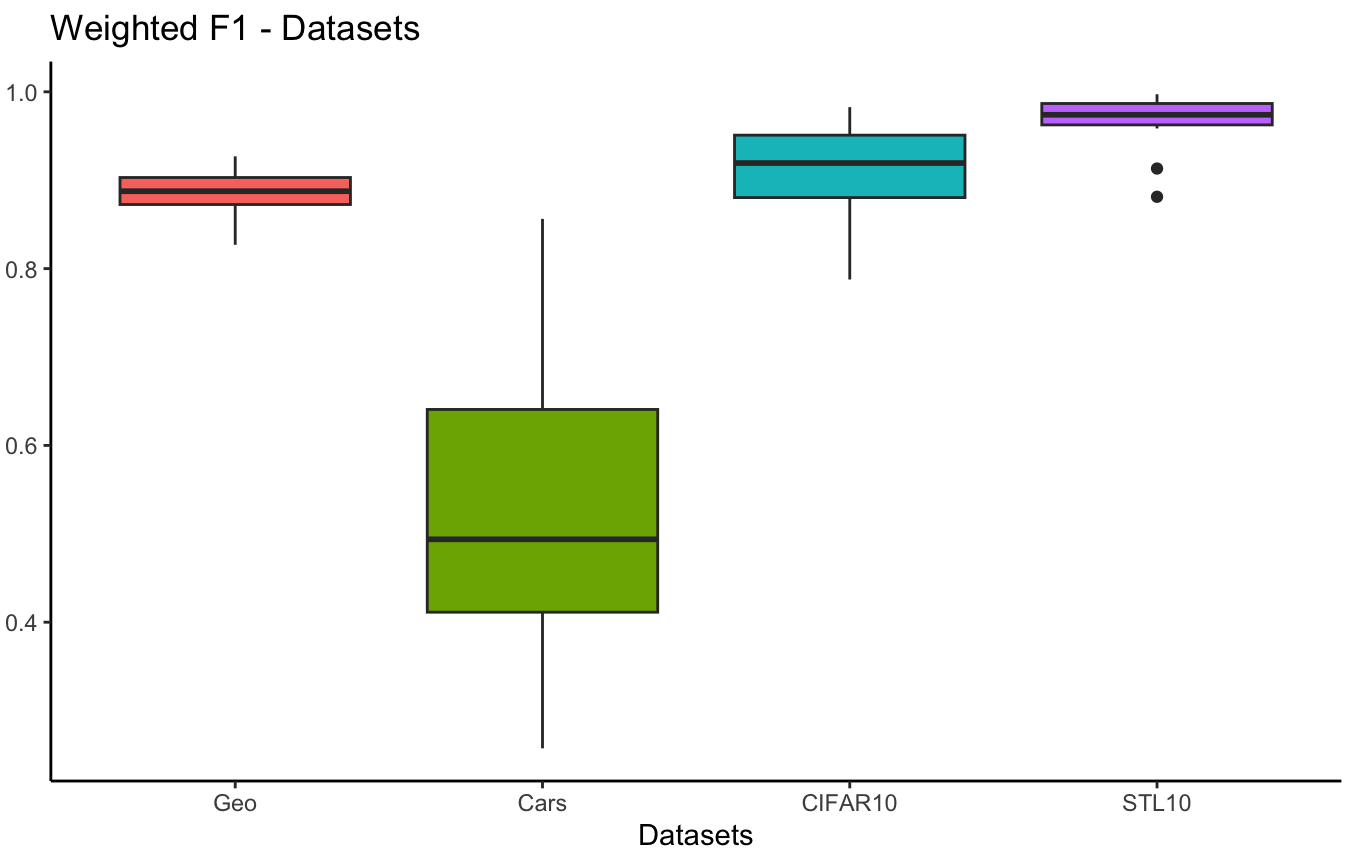}
\caption{Boxplot of weighted f1-score for each dataset.}
\label{fig:boxdatweifsco}
\end{figure}

The following boxplots represent the performances considering all the models in each dataset, according to different metrics. Figure \ref{fig:boxdatacc} represents the variation of accuracy. Figure \ref{fig:boxdatmacpre} shows the variation of macro precision. Figure \ref{fig:boxdatmacrec} indicates the variation of macro recall. Figure \ref{fig:boxdatmacfsco} demonstrates the variation of macro f1-score. Figure \ref{fig:boxdatweipre} presents the variation of weighted precision. Figure \ref{fig:boxdatweirec} represents the variation of weighted recall. Figure \ref{fig:boxdatweifsco} shows the variation of weighted f1-score in each dataset.

The boxplots represented in Figures \ref{fig:boxdatacc}-\ref{fig:boxdatweifsco} present a similar pattern across the different metrics. There are subtle differences when comparing the macro averages of precision, recall, and f-score with accuracy and the weighted averages of those metrics. These boxplots emphasize some properties found in the previous graphs (Figures \ref{fig:gldatacc} - \ref{fig:gldatweifsco}). Firstly, these charts provide evidence that the Stanford Cars dataset is the most challenging among those analyzed, with the worst performances and the greatest variability of performances in all metrics. Besides that, we can notice also that the STL10 dataset and Geological Images have a smaller variability in the performance of the different models when compared with the other two datasets.

\section{Conclusions} \label{sec:conclusion}

In this work, our goal was to compare the performance of sixteen pre-trained neural networks for FE in four different datasets, with an emphasis on geological image classification.

By analyzing the accuracy and macro and weighted averages of f-score, recall, and precision, our experiments have shown that CLIP-ViT-B achieved the best results for the Geological Images dataset, which was the main focus of this work. Our experiments showed also that CLIP-ResNet50 and VisionTransformer-H/14 also achieved similar performances.

When we focus on the general performances, considering all the datasets, our experiments suggest that CLIP-ViT-B and VisionTransformer-H/14 achieved better performance results for all datasets and low variability in their performances. Besides that, CLIP-ResNet50 achieved performance similar to the performance achieved by CLIP-ViT-B, and VisionTransformer-H/14 and even lower variability. It is important to notice that CLIP-ViT-B, VisionTransformer-H/14, and CLIP-ResNet50 include transformers in their architectures. Thus, our results suggest that the principles underlying the transformers can be the reason corroborating these remarkable results, but further studies should be carried out to investigate this hypothesis.

Among the CNN-based architectures, ConvNeXt Large presents the best performance, in general, and lower variability when compared to other CNN-based architectures. AlexNet showed the worst performance and high variability. Besides that, ResNeXt101-64x4D, Wide ResNet 101-2, and Inception V3 also showed high variability.

Our analysis also showed that the performances of models based solely on CNN architectures present a high Pearson correlation in all performance metrics. However, the performances of CLIP-ResNet50, CLIP-ViT-B, and VisionTransformer-H/14 models show a lower correlation with other models based solely on CNN. This difference in performance is hypothesized to be due to differences regarding the basic principles of the architecture of these models. However, the correlations among the performances of CLIP-ResNet50, CLIP-ViT-B, and VisionTransformer-H/14 models are not high, if compared to correlations among the performances of CNN-based models. Further studies are needed to better investigate this finding. 

Our analysis also has shown that the selected models performed better on the STL10 dataset, followed by CIFAR-10, then the Geological Images dataset, and finally the Stanford Cars dataset. Thus, we can conclude that the Stanford Cars dataset is the most challenging dataset evaluated in this work. The Stanford Cars present a particularly large image size (when compared with CIFAR-10 and STL10, for example) and a high amount of classes with just a few samples per class. These characteristics may explain this result. The Geological Images dataset shares some of the properties of the Stanford Cars dataset, but it presents a higher imbalance, has a lower amount of classes, and has more images per class, in general.

The investigation presented in this work can provide evidence that supports the choice of models for transfer learning in image classification tasks involving the Geological Images dataset. Since our evaluation also covered other image datasets, it can also suggest reasonable choices for transfer learning in other domains.

In future works, to make the analysis more comprehensive, it is important to expand it by including other image datasets. Besides that, the investigation can also be expanded to include more pre-trained models that eventually were not considered in the scope of this work. Furthermore, future works could also investigate the relationship between the underlying principles of each architecture, the properties of the datasets used in the pre-training of these models, and the properties of the target datasets, in which the pre-trained models are applied to extract features. This investigation can reveal insights into what makes the pre-trained model best suited for each task.

\bibliographystyle{unsrt}

\begin{thebibliography}{10}

\bibitem{pferd:10}
J.W. Pferd.
\newblock The challenges of integrating structured and unstructured data.
\newblock In {\em 14th Petroleum Network Education Conference}. s.n, S.l., 2010.

\bibitem{hollink:03}
Laura Hollink, Guus Schreiber, Jan Wielemaker, Bob Wielinga, et~al.
\newblock Semantic annotation of image collections.
\newblock In {\em Knowledge capture}, volume~2, 2003.

\bibitem{lanchantin:21}
Jack Lanchantin, Tianlu Wang, Vicente Ordonez, and Yanjun Qi.
\newblock General multi-label image classification with transformers.
\newblock In {\em Proceedings of the IEEE/CVF Conference on Computer Vision and Pattern Recognition (CVPR)}, pages 16478--16488, June 2021.

\bibitem{deng:09}
Jia Deng, Wei Dong, Richard Socher, Li-Jia Li, Kai Li, and Li~Fei-Fei.
\newblock Imagenet: A large-scale hierarchical image database.
\newblock In {\em 2009 IEEE conference on computer vision and pattern recognition}, pages 248--255. Ieee, 2009.

\bibitem{mallouh:19}
Arafat~Abu Mallouh, Zakariya Qawaqneh, and Buket~D Barkana.
\newblock Utilizing cnns and transfer learning of pre-trained models for age range classification from unconstrained face images.
\newblock {\em Image and Vision Computing}, 88:41--51, 2019.

\bibitem{arslan:21}
Yusuf Arslan, Kevin Allix, Lisa Veiber, Cedric Lothritz, Tegawend{\'e}~F Bissyand{\'e}, Jacques Klein, and Anne Goujon.
\newblock A comparison of pre-trained language models for multi-class text classification in the financial domain.
\newblock In {\em Companion Proceedings of the Web Conference 2021}, pages 260--268, 2021.

\bibitem{kieffer2017convolutional}
Brady Kieffer, Morteza Babaie, Shivam Kalra, and Hamid~R Tizhoosh.
\newblock Convolutional neural networks for histopathology image classification: Training vs. using pre-trained networks.
\newblock In {\em 2017 Seventh International Conference on Image Processing Theory, Tools and Applications (IPTA)}, pages 1--6. IEEE, 2017.

\bibitem{mormont2018comparison}
Romain Mormont, Pierre Geurts, and Rapha{\"e}l Mar{\'e}e.
\newblock Comparison of deep transfer learning strategies for digital pathology.
\newblock In {\em Proceedings of the IEEE conference on computer vision and pattern recognition workshops}, pages 2262--2271, 2018.

\bibitem{TodescatoGBC23}
Matheus~V. Todescato, Luan~Fonseca Garcia, Dennis~Giovani Balreira, and Joel~Lu{\'{\i}}s Carbonera.
\newblock Multiscale context features for geological image classification.
\newblock In Joaquim Filipe, Michal Smialek, Alexander Brodsky, and Slimane Hammoudi, editors, {\em Proceedings of the 25th International Conference on Enterprise Information Systems, {ICEIS} 2023, Volume 1, Prague, Czech Republic, April 24-26, 2023}, pages 407--418. {SCITEPRESS}, 2023.

\bibitem{abel2019knowledge}
Mara Abel, Eduardo Sim{\~o}es~Lopes Gastal, Cassiana Roberta~Lizzoni Michelin, Luiza~Gon{\c{c}}alves Maggi, Bruno~Eduardo Firnkes, Felix Eduardo~Huaroto Pachas, and Renata dos Santos~Alvarenga.
\newblock A knowledge organization system for image classification and retrieval in petroleum exploration domain.
\newblock In {\em ONTOBRAS}, 2019.

\bibitem{todescato2024multiscale}
Matheus~V Todescato, Luan~F Garcia, Dennis~G Balreira, and Joel~L Carbonera.
\newblock Multiscale patch-based feature graphs for image classification.
\newblock {\em Expert Systems with Applications}, 235:121116, 2024.

\bibitem{KrauseStarkDengFei:13}
Jonathan Krause, Michael Stark, Jia Deng, and Li~Fei-Fei.
\newblock 3d object representations for fine-grained categorization.
\newblock In {\em 4th International IEEE Workshop on 3D Representation and Recognition (3dRR-13)}, Sydney, Australia, 2013.

\bibitem{krizhevsky:09}
Alex Krizhevsky, Geoffrey Hinton, et~al.
\newblock Learning multiple layers of features from tiny images.
\newblock 2009.

\bibitem{coates:11}
Adam Coates, Andrew Ng, and Honglak Lee.
\newblock An analysis of single-layer networks in unsupervised feature learning.
\newblock In {\em Proceedings of the fourteenth international conference on artificial intelligence and statistics}, pages 215--223. JMLR Workshop and Conference Proceedings, 2011.

\bibitem{alzubaidi:21}
Laith Alzubaidi et~al.
\newblock Review of deep learning: Concepts, cnn architectures, challenges, applications, future directions.
\newblock {\em Journal of big Data}, 8(1):1--74, 2021.

\bibitem{dosovitskiy:20}
Alexey Dosovitskiy, Lucas Beyer, Alexander Kolesnikov, Dirk Weissenborn, Xiaohua Zhai, Thomas Unterthiner, Mostafa Dehghani, Matthias Minderer, Georg Heigold, Sylvain Gelly, et~al.
\newblock An image is worth 16x16 words: Transformers for image recognition at scale.
\newblock {\em arXiv preprint arXiv:2010.11929}, 2020.

\bibitem{maniar:18}
Hiren Maniar et~al.
\newblock Machine-learning methods in geoscience.
\newblock In {\em 2018 SEG International Exposition and Annual Meeting}. OnePetro, 2018.

\bibitem{karpatne:18}
Anuj Karpatne et~al.
\newblock Machine learning for the geosciences: Challenges and opportunities.
\newblock {\em IEEE Transactions on Knowledge and Data Engineering}, 31(8):1544--1554, 2018.

\bibitem{fawaz:18}
Hassan~Ismail Fawaz et~al.
\newblock Transfer learning for time series classification.
\newblock In {\em 2018 IEEE international conference on big data (Big Data)}, pages 1367--1376. IEEE p, 2018.

\bibitem{luminiAndnanni:19}
Alessandra Lumini and Loris Nanni.
\newblock Deep learning and transfer learning features for plankton classification.
\newblock {\em Ecological informatics}, 51:33--43, 2019.

\bibitem{kumarAndanuarAndhafizah:22}
Jayapalan~Senthil Kumar, Syahid Anuar, and Noor~Hafizah Hassan.
\newblock Transfer learning based performance comparison of the pre-trained deep neural networks.
\newblock {\em International Journal of Advanced Computer Science and Applications}, 13:1, 2022.

\bibitem{aboubAndzengelerAndhandmann:22}
Nermeen Abou~Baker, Nico Zengeler, and Handmann.
\newblock Uwe. a transfer learning evaluation of deep neural networks for image classification.
\newblock {\em Machine Learning and Knowledge Extraction}, 4(1):22--41, 2022.

\bibitem{delima:19}
Rafael~Pires De~Lima et~al.
\newblock Deep convolutional neural networks as a geological image classification tool.
\newblock {\em The Sedimentary Record}, 17(2):4--9, 2019.

\bibitem{sun:21}
Huiming Sun et~al.
\newblock Convolutional neural networks based remote sensing scene classification under clear and cloudy environments.
\newblock In {\em Proceedings of the IEEE/CVF International Conference on Computer Vision p}, pages 713--720, 2021.

\bibitem{chevitarese:18}
Daniel Chevitarese et~al.
\newblock {\em Transfer learning applied to seismic images classification}.
\newblock AAPG Annual and Exhibition, 2018.

\bibitem{chevitarese:18efficient}
Daniel~Salles Chevitarese, Daniela Szwarcman, Emilio~Vital Brazil, and Bianca Zadrozny.
\newblock Efficient classification of seismic textures.
\newblock In {\em 2018 International Joint Conference on Neural Networks (IJCNN)}, pages 1--8. IEEE, 2018.

\bibitem{cunha:20}
Augusto Cunha et~al.
\newblock Seismic fault detection in real data using transfer learning from a convolutional neural network pre-trained with synthetic seismic data.
\newblock {\em Computers \& Geosciences}, 135(10434):4, 2020.

\bibitem{radford:21}
Alec Radford, Jong~Wook Kim, Chris Hallacy, Aditya Ramesh, Gabriel Goh, Sandhini Agarwal, Girish Sastry, Amanda Askell, Pamela Mishkin, Jack Clark, et~al.
\newblock Learning transferable visual models from natural language supervision.
\newblock In {\em International conference on machine learning}, pages 8748--8763. PMLR, 2021.

\bibitem{krizhevsky:14}
Alex Krizhevsky.
\newblock One weird trick for parallelizing convolutional neural networks.
\newblock {\em arXiv preprint arXiv:1404.5997}, 2014.

\bibitem{he:16}
Kaiming He et~al.
\newblock Deep residual learning for image recognition.
\newblock In {\em Proceedings of the IEEE conference on computer vision and pattern recognition p}, pages 770--778, 2016.

\bibitem{liu:22}
Zhuang Liu, Hanzi Mao, Chao-Yuan Wu, Christoph Feichtenhofer, Trevor Darrell, and Saining Xie.
\newblock A convnet for the 2020s.
\newblock In {\em Proceedings of the IEEE/CVF Conference on Computer Vision and Pattern Recognition}, pages 11976--11986, 2022.

\bibitem{huang:17}
Gao Huang, Zhuang Liu, Laurens Van Der~Maaten, and Kilian~Q Weinberger.
\newblock Densely connected convolutional networks.
\newblock In {\em Proceedings of the IEEE conference on computer vision and pattern recognition}, pages 4700--4708, 2017.

\bibitem{szegedy:15}
Christian Szegedy, Wei Liu, Yangqing Jia, Pierre Sermanet, Scott Reed, Dragomir Anguelov, Dumitru Erhan, Vincent Vanhoucke, and Andrew Rabinovich.
\newblock Going deeper with convolutions.
\newblock In {\em Proceedings of the IEEE conference on computer vision and pattern recognition}, pages 1--9, 2015.

\bibitem{szegedy:16}
Christian Szegedy, Vincent Vanhoucke, Sergey Ioffe, Jon Shlens, and Zbigniew Wojna.
\newblock Rethinking the inception architecture for computer vision.
\newblock In {\em Proceedings of the IEEE conference on computer vision and pattern recognition}, pages 2818--2826, 2016.

\bibitem{tan:19}
Mingxing Tan, Bo~Chen, Ruoming Pang, Vijay Vasudevan, Mark Sandler, Andrew Howard, and Quoc~V Le.
\newblock Mnasnet: Platform-aware neural architecture search for mobile.
\newblock In {\em Proceedings of the IEEE/CVF conference on computer vision and pattern recognition}, pages 2820--2828, 2019.

\bibitem{howard:19}
Andrew Howard, Mark Sandler, Grace Chu, Liang-Chieh Chen, Bo~Chen, Mingxing Tan, Weijun Wang, Yukun Zhu, Ruoming Pang, Vijay Vasudevan, et~al.
\newblock Searching for mobilenetv3.
\newblock In {\em Proceedings of the IEEE/CVF international conference on computer vision}, pages 1314--1324, 2019.

\bibitem{radosavovic:20}
Ilija Radosavovic, Raj~Prateek Kosaraju, Ross Girshick, Kaiming He, and Piotr Doll{\'a}r.
\newblock Designing network design spaces.
\newblock In {\em Proceedings of the IEEE/CVF conference on computer vision and pattern recognition}, pages 10428--10436, 2020.

\bibitem{xie:17}
Saining Xie, Ross Girshick, Piotr Doll{\'a}r, Zhuowen Tu, and Kaiming He.
\newblock Aggregated residual transformations for deep neural networks.
\newblock In {\em Proceedings of the IEEE conference on computer vision and pattern recognition}, pages 1492--1500, 2017.

\bibitem{ma:18}
Ningning Ma, Xiangyu Zhang, Hai-Tao Zheng, and Jian Sun.
\newblock Shufflenet v2: Practical guidelines for efficient cnn architecture design.
\newblock In {\em Proceedings of the European conference on computer vision (ECCV)}, pages 116--131, 2018.

\bibitem{iandola:16}
Forrest~N Iandola, Song Han, Matthew~W Moskewicz, Khalid Ashraf, William~J Dally, and Kurt Keutzer.
\newblock Squeezenet: Alexnet-level accuracy with 50x fewer parameters and< 0.5 mb model size.
\newblock {\em arXiv preprint arXiv:1602.07360}, 2016.

\bibitem{simonyanAndzisserman:14}
Karen Simonyan and Andrew Zisserman.
\newblock Very deep convolutional networks for large-scale image recognition.
\newblock {\em arXiv preprint arXiv:1409.1556}, 2014.

\bibitem{zagoruyko:16}
Sergey Zagoruyko and Nikos Komodakis.
\newblock Wide residual networks.
\newblock {\em arXiv preprint arXiv:1605.07146}, 2016.

\bibitem{cohen:09}
Israel Cohen, Yiteng Huang, Jingdong Chen, Jacob Benesty, Jacob Benesty, Jingdong Chen, Yiteng Huang, and Israel Cohen.
\newblock Pearson correlation coefficient.
\newblock {\em Noise reduction in speech processing}, pages 1--4, 2009.

\end{thebibliography}

\end{document}